\documentclass[sigconf]{acmart}

\usepackage{multirow}
\usepackage{makecell}
\usepackage{adjustbox}
\usepackage{utfsym}
\definecolor{sisrcolor}{RGB}{235,224,234}
\definecolor{vsrcolor}{RGB}{195,232,209}
\definecolor{modelcolor}{RGB}{253,230,215}

\AtBeginDocument{%
  }

\copyrightyear{2025}
\acmYear{2025}
\setcopyright{acmlicensed}
\acmConference[MM '25] {Proceedings of the 33rd ACM International Conference on Multimedia}{October 27--31, 2025}{Dublin, Ireland.}
\acmBooktitle{Proceedings of the 33rd ACM International Conference on Multimedia (MM '25), October 27--31, 2025, Dublin, Ireland}
\acmISBN{979-8-4007-2035-2/2025/10}
\acmDOI{10.1145/3746027.3755117}

\begin{document}
\title{UltraVSR: Achieving Ultra-Realistic Video Super-Resolution with Efficient One-Step Diffusion Space}

        \author{Yong Liu}
        \affiliation{
          \institution{Xi'an Jiaotong University}
          \city{Xi'an}
          \country{China}
        }
        \email{liuy1996v@qq.com}
        
        \author{Jinshan Pan}
        \affiliation{
          \institution{Nanjing University of Science and Technology}
          \city{Nanjing}
          \country{China}
        }
        \email{sdluran@gmail.com}
        
        \author{Yinchuan Li}
        \affiliation{
          \institution{Huawei Noah's Ark Lab}
          \city{Beijing}
          \country{China}
        }

        \author{Qingji Dong}
        \affiliation{
          \institution{Xi'an Jiaotong University}
          \city{Xi'an}
          \country{China}
        }

        \author{Chao Zhu}
        \affiliation{
          \institution{Xi'an Jiaotong University}
          \city{Xi'an}
          \country{China}
        }

        \author{Yu Guo}
        \affiliation{
          \institution{Xi'an Jiaotong University}
          \city{Xi'an}
          \country{China}
        }

        \author{Fei Wang}
        \authornote{Corresponding author. Author affiliation: National Key Laboratory of Human-Machine Hybrid Augmented Intelligence, National Engineering Research Center of Visual Information and Applications, Institute of Artificial Intelligence and Robotics, Xi'an Jiaotong University.}
        \affiliation{
          \institution{Xi'an Jiaotong University}
          \city{Xi'an}
          \country{China}
        }  
        \email{wfx@xjtu.edu.cn}
        
\renewcommand{\shortauthors}{Yong Liu et al.}

\begin{abstract}
Diffusion models have shown great potential in generating realistic image detail. 
However, adapting these models to video super-resolution (VSR) remains challenging due to their inherent stochasticity and lack of temporal modeling. 
Previous methods have attempted to mitigate this issue by incorporating motion information and temporal layers. 
However, unreliable motion estimation from low-resolution videos and costly multiple sampling steps with deep temporal layers limit them to short sequences. 
In this paper, we propose UltraVSR, a novel framework that enables ultra-realistic and temporally-coherent VSR through an efficient one-step diffusion space. 
A central component of UltraVSR is the Degradation-aware Reconstruction Scheduling (DRS), which estimates a degradation factor from the low-resolution input and transforms the iterative denoising process into a single-step reconstruction from low-resolution to high-resolution videos. 
To ensure temporal consistency, we propose a lightweight Recurrent Temporal Shift (RTS) module, including an RTS-convolution unit and an RTS-attention unit.
By partially shifting feature components along the temporal dimension, it enables effective propagation, fusion, and alignment across frames without explicit temporal layers. 
The RTS module is integrated into a pretrained text-to-image diffusion model and is further enhanced through Spatio-temporal Joint Distillation (SJD), which improves temporally coherence while preserving realistic details. 
Additionally, we introduce a Temporally Asynchronous Inference (TAI) strategy to capture long-range temporal dependencies under limited memory constraints. 
Extensive experiments show that UltraVSR achieves state-of-the-art performance, both qualitatively and quantitatively, in a single sampling step. 
Code is available at \url{https://github.com/yongliuy/UltraVSR}.
\end{abstract}


\begin{teaserfigure}
  \centering
  \includegraphics[width=0.95\textwidth]{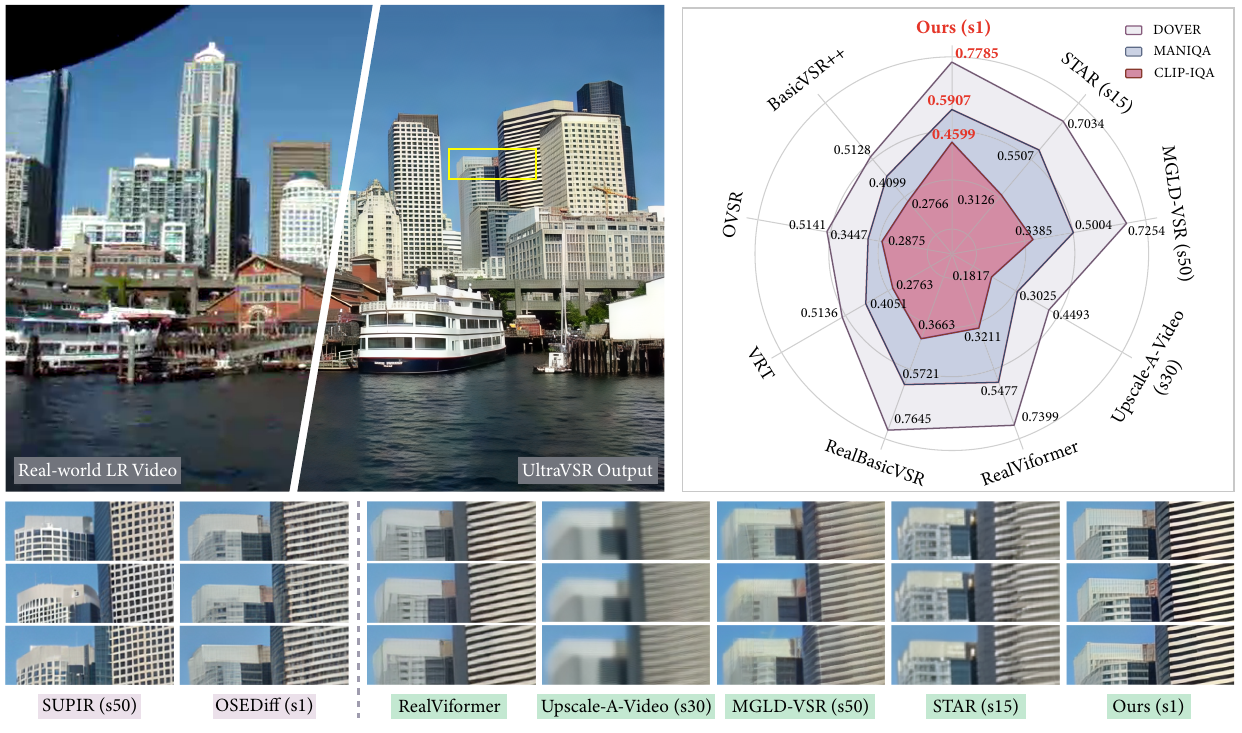}
  \vspace{-1.5em}
  \caption{Quantitative and qualitative VSR comparisons. Top-left: An example on a real-world low-resolution (LR) video frame; 
  Top-right: Quantitative comparison with state-of-the-art \raisebox{0pt}[1.5ex][0.5ex]{\colorbox{vsrcolor}{VSR approaches}} on the VideoLQ benchmark~\cite{chan2022investigating}; 
  Bottom: Visual comparison against both \raisebox{0pt}[1.5ex][0.5ex]{\colorbox{sisrcolor}{SISR approaches}} and \raisebox{0pt}[1.5ex][0.5ex]{\colorbox{vsrcolor}{VSR approaches}} across three consecutive frames. 
  Our one-step (s1) approach achieves leading scores across all three metrics with visually realistic and temporally coherent results.}
  \label{fig:teaser}
\end{teaserfigure}

\maketitle

\section{Introduction}
Video super-resolution (VSR) aims to reconstruct high-resolution (HR) videos from low-resolution (LR) inputs, a task that has garnered significant attention in multimedia and computer vision communities. 
Unlike single-image super-resolution (SISR), VSR presents additional challenges due to the necessity of maintaining temporal consistency across frames while recovering high-frequency details. 
With the rapid advancement of deep learning, numerous VSR approaches~\cite{chan2021basicvsr, chan2022basicvsr++, fuoli2019efficient, gupta2021ada, jin2023kernel, leng2022icnet, xiao2024asymmetric, ai2024skipvsr} have emerged over the past decade. 
However, these methods still struggle to restore realistic textures under complex and unknown real-world degradations. 

\begin{figure}
  \centering
  \includegraphics[width=0.32\textwidth]{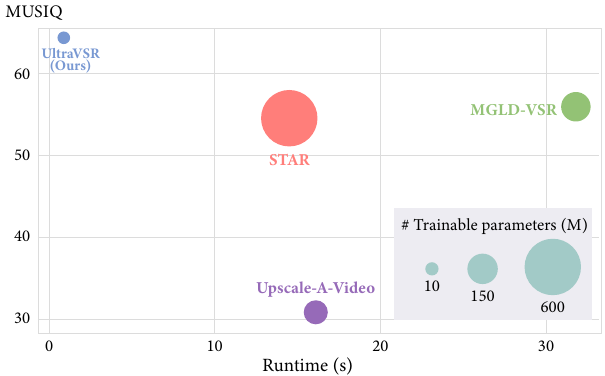}
  \vspace{-1em}
  \caption{Comparison of runtime and metrics among diffusion-based VSR approaches. Note that the runtime is measured with the HR sequence upscaled to 720$\times$1280. Our approach achieves the fastest speed with better quality.}
  \label{fig:runtime_against_metrics}
  \vspace{-2em}
\end{figure}

Diffusion models~\cite{ho2020denoising, sohl2015deep} have achieved remarkable success in image synthesis~\cite{rombach2022high, dhariwal2021diffusion}, image editing~\cite{kawar2023imagic, yang2023paint} and SISR~\cite{wang2023exploring, wang2024sinsr}. 
However, adapting these models to VSR remains challenging due to the lack of temporal modeling capabilities. 
Additionally, the inherent randomness of the diffusion process further exacerbates temporal artifacts such as flickering across frames, e.g., the results of SUPIR~\cite{yu2024scaling} and OSEDiff~\cite{wu2024one} in Fig.~\ref{fig:teaser}. 
Recent approaches, such as MGLD-VSR~\cite{yang2025motion} and Upscale-A-Video~\cite{zhou2024upscale}, have attempted to improve temporal consistency by integrating 3D convolutions, temporal attention mechanisms, and motion propagation modules into image diffusion models. 
Alternatively, text-to-video diffusion models have also been explored for VSR~\cite{xie2025star}. 
Nevertheless, these approaches suffer from high computational costs due to the use of temporal layers and unreliable motion information estimated from LR videos. 
Moreover, these approaches require a large number of sampling steps for each frame, making inference inefficient and impractical, especially for memory-constrained processing of high-resolution and long-duration video sequences.

This paper proposes UltraVSR, an efficient one-step diffusion model for real-world VSR.  
In contrast to multi-step VSR approaches, UltraVSR introduces a Degradation-aware Reconstruction Scheduling (DRS) that reformulates the conventional multi-step iterative denoising paradigm as a single-step reconstruction process.
DRS first estimates a degradation factor from the input LR video and then maps it with the diffusion noise schedule, thereby achieving single-step reconstruction from LR input to HR video. 
This scheme eliminates the randomness in the diffusion process and significantly improves computational efficiency. 
To maintain temporal consistency across frames, we develop a lightweight yet effective Recurrent Temporal Shift (RTS) module, which enables efficient inter-frame feature propagation, fusion, and alignment through two novel units: 
(1) RTS-convolution unit, which partitions feature maps by channels and shifts groups temporally across adjacent frames for localized feature alignment using 2D convolution; 
(2) RTS-attention unit, which incorporates temporal shifting operations on $Key$/$Value$ tensors prior to self-attention computation, allowing the model to capture global contextual dependencies.

During training, the RTS module is embedded into a pretrained text-to-image diffusion model and subsequently optimized through Spatio-temporal Joint Distillation (SJD). 
SJD introduces dual temporal regularizers to separately model real and synthetic video distributions. 
By minimizing both spatial and temporal gradient discrepancies between two distributions, SJD effectively guides the one-step reconstruction process toward producing visually realistic and temporally consistent HR videos. 
At inference, we further propose a memory-efficient Temporally Asynchronous Inference (TAI) strategy that decouples spatial and temporal processing by segmenting long video sequences into mini-batches, which are processed independently and then merged prior to RTS propagation. 
In this way, TAI significantly reduces memory overhead while enabling the model to capture long-range temporal dependencies. 
Experiments demonstrate that UltraVSR achieves better performance with significantly faster inference, as shown in Fig.~\ref{fig:teaser} and Fig.~\ref{fig:runtime_against_metrics}. 

In summary, the main contributions of this paper are as follows: 
\begin{itemize}
	\item We propose UltraVSR, an efficient one-step diffusion framework by introducing a Degradation-aware Reconstruction Scheduling (DRS) for high-quality video reconstruction. To the best of our knowledge, this is the first work to tackle the VSR problem in just one sampling step. 
	\item We develop a lightweight Recurrent Temporal Shift (RTS) that enables effective feature propagation, fusion, and alignment across frames by partially shifting feature components temporally, without relying on explicit temporal layers. 
	\item We introduce Spatio-temporal Joint Distillation (SJD) to enhance temporally coherence and content realism, and propose Temporally Asynchronous Inference (TAI) for memory-efficient long-range dependency modeling. 
	\item Extensive experiments show that UltraVSR outperforms state-of-the-art methods in both visual quality and temporal consistency, while significantly accelerating inference speed. 
\end{itemize}

\section{Related Work}
\label{sec:related}

\subsection{Real-World VSR}
Numerous VSR methods~\cite{chan2021basicvsr, fuoli2019efficient, gupta2021ada, li2023simple, liang2022recurrent, xiao2020space, leng2022icnet} have been developed to enhance video quality under real-world degradation. 
For instance, Chan \textit{et al.} proposed RealBasicVSR~\cite{chan2022investigating}, which removes degradations in videos by introducing a pre-cleaning module and a stochastic degradation scheme. 
More recently, 
Zhang \textit{et al.} introduced a channel-attention-based recurrent model, RealViformer~\cite{zhang2024realviformer}, based on findings from investigating channel and spatial attention in a real-world setting. 
Despite these advancements, many approaches remain sensitive to complex degradations due to limited modeling capacity, leading to poor generalization.

\subsection{Diffusion Model-based VSR}
Diffusion models~\cite{ho2020denoising, sohl2015deep} have demonstrated strong performance in generating realistic details~\cite{rombach2022high, kawar2023imagic, li2025generative, yang2023paint, yang2024cogvideox, blattmann2023align}. 
Recent studies have attempted to scale up diffusion models for the VSR task in two ways. 
On one hand, Zhou \textit{et al.}\cite{zhou2024upscale} and Yang \textit{et al.}\cite{yang2025motion} leverage image diffusion models and propose Upscale-A-Video and MGLD-VSR, respectively, by incorporating additional temporal layers and motion information. 
On the other hand, Xie \textit{et al.}\cite{xie2025star} build upon a pretrained text-to-video diffusion model and introduce STAR, which employs ControlNet\cite{zhang2023adding} to inject LR video information into the generation process. 
However, these approaches often rely on multiple sampling steps and complex temporal layers, resulting in high inference latency. 
To overcome these challenges, we propose an efficient one-step diffusion model for real-world VSR. 

\section{Proposed Method}
\label{sec:method}

\subsection{Overview}
\begin{figure*}[t]
    \setlength{\abovecaptionskip}{0pt}
    \setlength{\belowcaptionskip}{-1pt}
    \centering
    \includegraphics[width=0.85\linewidth]{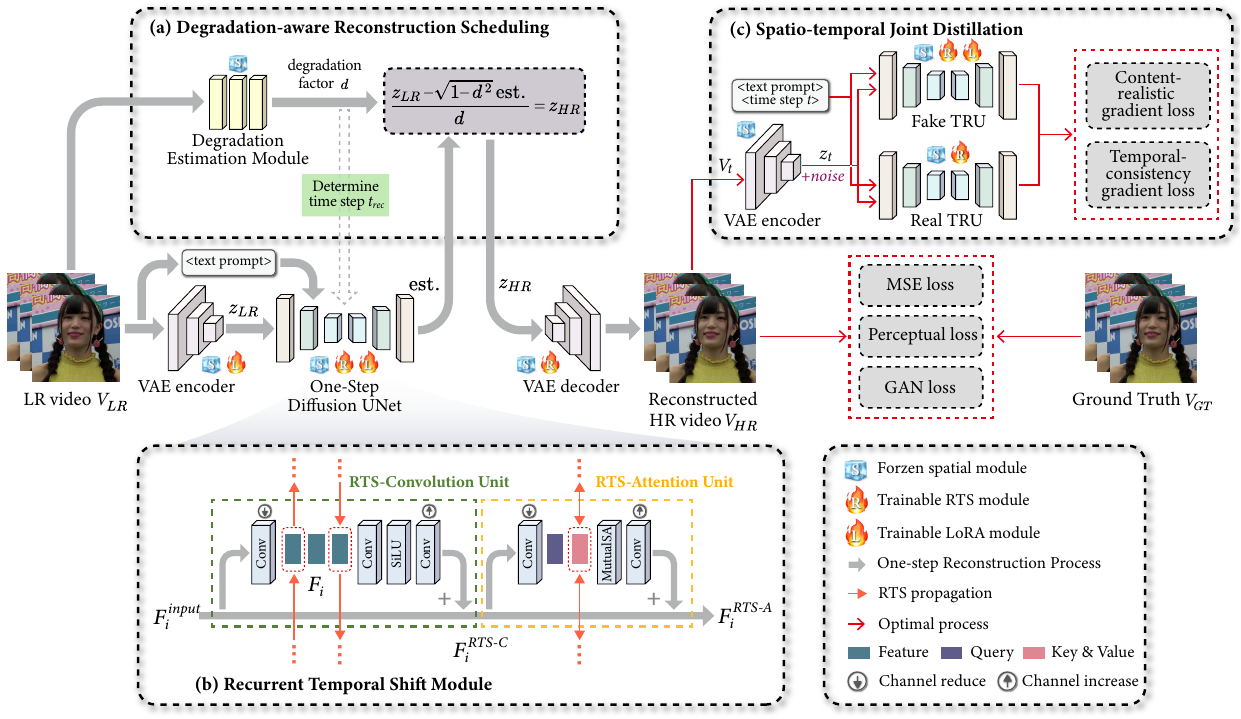}
    \caption{Training framework of the proposed UltraVSR. }
    \label{fig:network}
    \vspace{-1em}
\end{figure*}

Our UltraVSR is built upon the pretrained Stable Diffusion model~\cite{rombach2022high}, comprising a VAE encoder/decoder, a text encoder, and a diffusion UNet, as illustrated in Fig.\ref{fig:network}. 
To construct an efficient one-step reconstruction model $G_{\theta}$ to generate HR video $V_{HR}$ from LR input $V_{LR}$, we introduce three key components: 
(i) Degradation-aware Reconstruction Scheduling (DRS): Defines a one-step reconstruction from LR videos to HR videos (Sec.~\ref{DRS}); 
(ii) Recurrent Temporal Shift (RTS) Module: Incorporates temporal dependencies into the diffusion model by shifting specific feature components across frames, without relying on explicit temporal layers (Sec.~\ref{RTS}); 
(iii) Spatio-temporal Joint Distillation (SJD): Guides training with the content realistic and temporal coherent supervision (Sec.~\ref{SJD}). 
Given an input LR video sequence $V_{LR}$, UltraVSR first estimates a degradation factor $d$ and determines a corresponding reconstruction time step $t_{rec}$. 
Concurrently, a text prompt is simultaneously extracted from $V_{LR}$ via the text encoder, while a latent representation $ z_{LR} $ is derived using the VAE encoder. 
These variables are fed into the diffusion UNet together to produce intermediate estimates $est.$, which are subsequently refined through DRS to obtain the HR latent representation $ z_{HR} $. 
Finally, $ z_{HR}$ is passed to the VAE decoder to produce the high-quality HR output $V_{HR}$. 

During training, we incorporate LoRA layers~\cite{hu2022lora} into both the VAE encoder and diffusion UNet, achieving efficient parameter tuning while preserving generation priors. 
To ensure temporal consistency throughout the reconstruction process from $V_{LR}$ to $V_{HR}$, we integrate the RTS module across multiple scales of both the diffusion UNet and VAE decoder, components identified as primary sources of temporal flickering artifacts in the original architecture~\cite{yang2025motion,zhou2024upscale}. 
SJD is further used to enforce that the one-step reconstruction model generates $V_{HR}$ with both spatial realism and temporally coherence. 
The detailed network architectures are provided in the supplementary materials. 
We also propose a Temporally Asynchronous Inference (TAI) strategy for memory-efficient long-range dependency modeling in Sec.~\ref{TAI}.

\subsection{Degradation-aware Reconstruction Scheduling}
\label{DRS}

Standard diffusion models (DDPMs)~\cite{ho2020denoising, sohl2015deep} assume a forward diffusion process that gradually adds noise to real data $ \boldsymbol{z}_{0} $ following a fixed noise schedule: 
\begin{equation} 
	\boldsymbol{z}_{t}=\sqrt{\bar{\alpha}_{t}} \boldsymbol{z}_{0}+\sqrt{1-\bar{\alpha}_{t}} \boldsymbol{\epsilon}.  
	\label{eq_diffusion2}
\end{equation} 
where $t \in \{0, 1, 2, .., T\}$, $\bar{\alpha}_{t}=\prod_{i=0}^{t}\alpha_{t}$ is a hyperparameter controlling the noise strength. 
The reverse diffusion process iteratively denoises from $\boldsymbol{z}{T}$ to $\boldsymbol{z}{0}$, but this inherently results in computationally expensive inference.

Inspired by the consistency models~\cite{song2023consistency} and one-step generation models~\cite{yin2024one, nguyen2024swiftbrush, wang2024hero} that map the point $\mathbf{z}_{t}$ on the diffusion trajectory to the starting point $\mathbf{z}_{0}$ directly, we introduce DRS to enable a one-step reconstruction from LR latent $\mathbf{z}_{LR}$ to HR latent $\mathbf{z}_{HR}$. 
As shown in Fig.~\ref{fig:network} (a), we first estimate a degradation factor $d \in [0,1]$ to quantify the degradation severity of $V_{LR}$: 
\begin{equation}
	d=DEM(V_{LR}), 
    \label{rec2}
\end{equation} 
where the degradation estimation module ($DEM$) is implemented using the pretrained CLIP model~\cite{wang2023exploring}, which is a widely used no-reference quality assessment model. 
Here, $d=0$ corresponds to maximal video degradation, while $d=1$ indicates minimal degradation. 
This is consistent with the diffusion process described in Eq.~\ref{eq_diffusion2}: a lower $\bar{\alpha}_t$ pushes $\boldsymbol{z}_t$ closer to Gaussian noise, whereas a higher $\bar{\alpha}_t$ retains stronger fidelity to the real data distribution. 
Therefore, we formulate a direct mapping between the degradation factor $d$ and the noise schedule by defining $d=\sqrt{\bar{\alpha}_{t}}$, mathematically linking perceptual quality to the denoising process. 
In this way, the HR latent representation $\mathbf{z}_{HR}$ can be derived as: 
\begin{equation}
	\mathbf{z}_{HR}=\frac{\boldsymbol{z}_{LR}-\sqrt{1-d^2} est.}{d},  
    \label{rec3}
\end{equation} 
where $est.$ denotes the diffusion UNet's output.

Additionally, we determine the reconstruction time step $t_{rec}$ through nearest-neighbor matching against the predefined noise schedule $\{\sqrt{\bar{\alpha}_t}\}_{t=1}^T$: 
\begin{equation}
	t_{rec}=\text{arg}~\underset{t}{\text{min}} \left | d - \sqrt{\bar{\alpha}_t} \right |,
\end{equation} 
selecting the time step where $\sqrt{\bar{\alpha}_{t_{rec}}}$ most closely approximates the degradation factor $d$. 
This mechanism enables efficient one-step reconstruction by leveraging the pretrained generative prior.

\subsection{Recurrent Temporal Shift Module}
\label{RTS}
The key to VSR lies in effectively aggregating complementary spatio-temporal information from misaligned frames. 
To this end, we propose a lightweight yet effective Recurrent Temporal Shift (RTS) module that captures spatio-temporal dependencies without relying on computationally expensive temporal layers. 
As illustrated in Fig.\ref{fig:network} (b), the RTS module performs temporal feature propagation by bidirectionally shifting partitioned features across adjacent frames in a parameter-free manner, inspired by\cite{lin2019tsm, cao2023masactrl, munoz2021temporal}. 
It consists of two core components: the RTS-Convolution Unit and the RTS-Attention Unit, which collaborate to enable efficient information exchange across neighboring frames. 
To improve computational efficiency, we apply channel-reducing and channel-increasing convolutions at the beginning and end of each unit, respectively. 

\subsubsection{RTS-Convolution Unit.} 
Given an intermediate feature map $ F_i = Conv(F_i^{input}) \in \mathbb{R}^{C \times h \times w}$ for the $i$-th frame, RTS divides it into three equal channel-wise segments: $ F_i^{-1} $, $ F_i^{0} $, and $ F_i^{1} $, where each segment $ F_i^j \in \mathbb{R}^{T \times \frac{c}{3} \times h \times w} $. 
These segments are then shifted along the temporal dimension with offsets of +1, 0, and -1, respectively, where a positive offset denotes a forward shift, a negative offset denotes a backward shift, and zero implies no shift. 
The aggregated features for the $i$-th frame are then obtained by concatenating the temporally shifted feature segments: 
\begin{equation}
	F_i^{Agg}=Concat(F_{i-1}^{-1}, F_{i}^{0}, F_{i+1}^{1}). 
\end{equation} 
This strategy ensures that each frame receives contextual information from both past and future frames. 
To maintain consistent dimensions throughout the video, zero-padding is applied at the sequence boundaries. 
The aggregated features are processed via convolutional layers with ReLU activation followed by a residual connection to perform local temporal alignment: 
\begin{equation}
	F_i^{RTS-C} = Conv(ReLU(Conv(F_i^{Agg}))) + F_{i}^{input}. 
\end{equation} 
\subsubsection{RTS-Attention Unit.} 
Building on the observation that attention-based feature alignment inherently captures strong semantic priors in diffusion models~\cite{cao2023masactrl, alaluf2024cross, geyer2023tokenflow, tang2023emergent}, we further develop an RTS-attention unit to aggregate high-level global contextual dependencies across frames. 
Given the $Query$, $Key$, and $Value$ embeddings $\{Q_i, K_i, V_i\}=Conv(F_i^{RTS-C})$ for the $i$-th frame, RTS shifts $K_i$ and $V_i$ forward and backward temporally, allowing the model to query semantically correlated structures and textures from adjacent frames. 
We adopt Mutual Self-Attention (MutualSA)~\cite{cao2023masactrl} to aggregate semantic information across frames: 
\begin{equation}
	F_i^{MSA} = \sum_{j=-1}^{1} \text{MutualSA}(Q_i,K_{i+j},V_{i+j}), 
\end{equation} 
where $ \text{MutualSA}(Q_i,K_{i+j},V_{i+j}) =  \text{Softmax}(\frac{Q_i,K_{i+j}^T}{\sqrt{r}})V_{i+j} $, $r$ denotes the embedding dimension. 
The final output of the RTS-Attention Unit is obtained by:
\begin{equation}
	F_i^{RTS-A} = Conv(F_i^{MSA}) + F_i^{RTS-C}. 
\end{equation} 
Unlike conventional spatial self-attention, which operates independently on each frame, the RTS-attention unit aligns semantic information across the temporal dimension, thereby enhancing the temporal consistency of the reconstructed video.

\subsection{Spatio-temporal Joint Distillation}
\label{SJD}
To guide the proposed one-step reconstruction in generating visually realistic and temporally consistent HR videos, we introduce a dual-constrained optimization strategy, termed Spatio-temporal Joint Distillation (SJD). 
Our SJD shares motivation with other distribution matching generative models, such as SDS~\cite{poole2022dreamfusion} and VSD~\cite{yin2024one, wang2023prolificdreamer}, but extends these paradigms from the image domain to video, explicitly modeling inter-frame temporal consistency dependencies. 

As illustrated in Fig.\ref{fig:network} (c), SJD introduces two Temporal Regularized UNets (TRU), a real TRU $\mathcal{T}_{\text{real}}$ and a fake TRU $\mathcal{T}_{\text{fake}}$. 
They are derived by integrating our proposed RTS module into the Stable Diffusion model~\cite{rombach2022high}. 
During the training of the one-step reconstruction model $G_{\theta}$, $\mathcal{T}_{\text{real}}$ and $\mathcal{T}_{\text{fake}}$ are dynamically updated using ground-truth video $V_{GT}$ and the synthesized output $V_{HR}$, respectively. 
To capture both spatial realism and temporal consistency, we convert the noise predictions into video distributions using Eq.~\ref{eq_diffusion2} at the output of the TRUs. 
Conceptually, $\mathcal{T}_{\text{real}}$ can be interpreted as providing gradient directions that encourage the reconstruction to become more realistic and temporally coherent. 
In contrast, $\mathcal{T}_{\text{fake}}$ yields gradients that reflect undesirable characteristics such as visual artifacts and temporal flickering. 
To this end, we formulate the gradient update rule for $G_{\theta}$ by integrating a content-realistic gradient loss and a temporal-consistency gradient loss: 
\begin{equation}
\nabla\mathcal{L}_{\text{SJD}} = \nabla\mathcal{L}_{\text{realistic}} + \lambda \nabla\mathcal{L}_{\text{consistency}},
\end{equation}
where $\lambda$ balances content realism and temporal coherence of $V_{HR}$. 
This encourages the synthesized $V_{HR}$ toward higher perceptual realism and temporally coherence while suppressing visual artifacts and flickering. 
The content-realistic gradient loss is formulated as: 
\begin{equation}
\nabla\mathcal{L}_\text{realistic}= \mathbb{E}\left[\omega(t)\left(\mathcal{T}_{\text{real}}\left(z_{t}, t\right)-\mathcal{T}_{\text{fake}}\left(z_{t}, t\right)\right) \frac{d G}{d \theta} \right], 
\label{dmd}
\end{equation} 
where $z_t=\sqrt{\bar{\alpha}_{t}} z_{0}+\sqrt{1-\bar{\alpha}_{t}} \boldsymbol{\epsilon}$, $ z_{0} = \mathcal{E}(V_{HR}) $, $V_{HR}=G_{\theta}(V_{LR})$, $\boldsymbol{\epsilon} \sim \mathcal{N}(0, \mathbf{I})$, $\mathcal{E}$ denotes the pretrained VAE encoder, 
$\omega(t)$ is a weighting coefficient to normalize the gradient magnitudes across varying noise levels~\cite{yin2024one}. 
The temporal-consistency gradient loss enhances temporally coherence by penalizing discrepancies in the inter-frame gradients of two TRUs: 
\begin{equation}
\begin{aligned}
    &\nabla\mathcal{L}_\text{consistency} = 
    \mathbb{E}
    \Bigg[
    \omega(t) \Big(
    (\mathcal{T}_{\text{real}}(z_{t}, t)[:-1]
    - \mathcal{T}_{\text{real}}(z_{t}, t)[1:]) \\
    & - (\mathcal{T}_{\text{fake}}(z_{t}, t)[:-1]
    - \mathcal{T}_{\text{fake}}(z_{t}, t)[1:]) 
    \Big) \frac{d G}{d \theta} 
    \Bigg].
\end{aligned}
\end{equation} 

\subsection{Temporally Asynchronous Inference}
\label{TAI}
\begin{figure}[t]
    \setlength{\abovecaptionskip}{0pt}
    \setlength{\belowcaptionskip}{-1pt}
    \centering
    \includegraphics[width=0.85\linewidth]{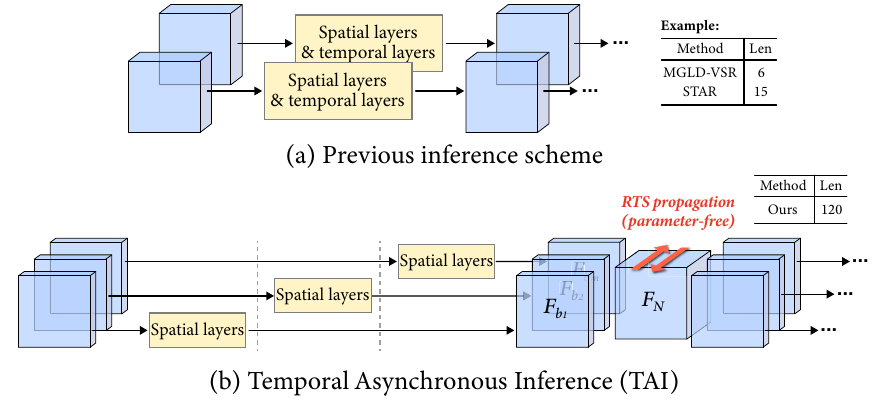}
    \caption{Previous inference scheme vs. the proposed TAI.}
    \label{fig:tai}
    \vspace{-1.3em}
\end{figure}

Existing diffusion-based VSR methods exhibit critical dependencies on temporal modules, incurring prohibitive memory overheads. 
As a result, such models require video segmentation into short clips for piecewise processing, as shown in Fig.~\ref{fig:tai} (a). 
For instance, under 48GB GPU memory constraints (NVIDIA RTX 8000), MGLD-VSR~\cite{yang2025motion} processes $\le $6 frames while STAR~\cite{xie2025star} handles $\le $15 frames for 2K (1440$\times$2560) reconstruction. 
This fragmentation disrupts temporal continuity and fundamentally limits long-range dependency modeling. 
UltraVSR circumvents this limitation through implicit temporal modeling via RTS propagation, eliminating explicit temporal layers. 
Building on this architectural innovation, we propose a Temporally Asynchronous Inference (TAI) strategy that enables efficient capture of long-range temporal dependencies.

As illustrated in Fig.~\ref{fig:tai} (b), given an $N$-frames low-resolution video $V_{LR}$, TAI first divides it into $m$ non-overlapping sub-sequences $\{V_{LR_1}, V_{LR_2}, \dots, V_{LR_m}\}$, each containing $b$ frames. 
Then, these sub-sequences are processed sequentially through 2D spatial layers to obtain feature maps $\{F_{b_1}, F_{b_2},...F_{b_m}\}$, drastically reducing peak memory consumption. 
At the beginning of the RTS propagation, all these features are concatenated into a unified feature tensor $F_N$, which is then augmented by bidirectional RTS propagation to achieve efficient temporal information exchange between frames. 
The temporally shifted features are then divided back into sub-sequences for further refinement. 

As a result, TAI enables memory-efficient inference while maintaining long-term temporal dependencies, capable of processing up to 120 frames at 1440$\times$2560 resolution identical hardware constraints, an \textbf{8$\times$} throughput improvement over STAR~\cite{xie2025star}. 
This breakthrough makes UltraVSR particularly suitable for high-quality reconstruction of long-duration video sequences under resource-limited environments.

\begin{table*}[!t] 
\centering
\caption{Quantitative comparsion with state-of-the-art VSR approaches on both synthetic and real-world benchmarks. The best and second best results are \textcolor{red!80!black}{\textbf{highlighted}} and \underline{underlined}, respectively.}
\vspace{-1.0em}
\label{table:Quantitative}
\begin{adjustbox}{width=0.90\linewidth}
\begin{tabular}{c|c|c c c c c c c c|c}
\hline
Datasets & Metrics & BasicVSR++~\cite{chan2022basicvsr++} & OVSR~\cite{yi2021omniscient} & RealBasicVSR~\cite{chan2022investigating} & 
VRT~\cite{liang2024vrt} &
RealViformer~\cite{zhang2024realviformer} & 
Upscale-A-Video~\cite{zhou2024upscale} & MGLD-VSR~\cite{shi2024motion}  & STAR~\cite{xie2025star} & Ours   \\ 
\hline
\hline
\multirow{6}{*}{\makecell[c]{REDS4}} 
&PSNR    & \textcolor{red!80!black}{\textbf{33.77}}  & 23.55 & 28.59 & \underline{32.97} & 28.18  & 25.62 & 27.49 & 24.08  & 24.50  \\    
&SSIM    & \textcolor{red!80!black}{\textbf{0.9173}} & 0.6743 & 0.8066 & \underline{0.9007} & 0.7896 & 0.6761 & 0.7783 & 0.6721 & 0.6962 \\ 
&MUSIQ   & 67.00  & 54.44 & \underline{67.03} & 65.68 & 64.58  &  32.30  & 58.94 & 60.94  & \textcolor{red!80!black}{\textbf{70.07}}  \\    
&CLIP-IQA& 0.3149 & 0.3383 & 0.3724 & \underline{0.3743} & 0.3558 &  0.1843 & 0.2908 & 0.3047 & \textcolor{red!80!black}{\textbf{0.4397}} \\
&MANIQA  & \underline{0.6752} & 0.5887 & 0.6671 & 0.6699 & 0.6545 & 0.3138 & 0.5837 & 0.6123 & \textcolor{red!80!black}{\textbf{0.6821}} \\ 
&DOVER  & 0.7182 & 0.6620 & 0.7209 & 0.7067 & 0.7152 & 0.4053 & 0.6717 & \underline{0.7227} & \textcolor{red!80!black}{\textbf{0.7304}} \\

\hline
\multirow{6}{*}{\makecell[c]{SPMCS}} 
&PSNR    & 23.39  & \textcolor{red!80!black}{\textbf{29.87}} & 22.87 & 23.43 & 23.45 & \underline{24.10}   & 23.43  & 22.95  & 23.08 \\    
&SSIM    & 0.6516 & \textcolor{red!80!black}{\textbf{0.8702}} & 0.6087 & \underline{0.6543} & 0.6153 & 0.6386  & 0.6372 & 0.6001 & 0.6202 \\ 
&MUSIQ   & 62.57  & 57.10 & \underline{69.44} & 63.44 & 66.10 & 56.31  & 64.08 & 55.14  & \textcolor{red!80!black}{\textbf{69.92}} \\    
&CLIP-IQA& 0.4341 & 0.4597 & \underline{0.4838} & 0.4388 & 0.4432  &  0.3353  & 0.4685 & 0.3888 & \textcolor{red!80!black}{\textbf{0.5377}} \\
&MANIQA  & 0.6263 & 0.6057 & \underline{0.6282} & 0.6134 & 0.6148 & 0.4798  & 0.5562 & 0.5640 & \textcolor{red!80!black}{\textbf{0.6386}} \\ 
&DOVER  & 0.6711 & 0.6031 & \underline{0.7506} & 0.6404 & 0.7206 & 0.5675  & 0.6431 & 0.6404 & \textcolor{red!80!black}{\textbf{0.7616}} \\

\hline
\multirow{6}{*}{\makecell[c]{UDM10}} 
&PSNR    & \underline{34.68} & 27.12 & 31.31 & \textcolor{red!80!black}{\textbf{36.03}} & 31.84  &  29.48  & 29.50  & 27.93  & 27.96  \\    
&SSIM    & \underline{0.9406} & 0.8318 & 0.8936 & \textcolor{red!80!black}{\textbf{0.9501}} & 0.9074 & 0.8458 & 0.8744 & 0.8284 & 0.8346 \\ 
&MUSIQ   & 63.26  & 57.94 & \underline{66.01} & 61.38 & 61.86  & 39.48  & 64.54 & 57.34  & \textcolor{red!80!black}{\textbf{66.46}} \\    
&CLIP-IQA& \textcolor{red!80!black}{\textbf{0.5091}} & 0.4338 & 0.4903 & 0.4697 & 0.4356  & 0.2356 & \underline{0.5083} & 0.3843 & 0.4818 \\
&MANIQA  & 0.5729 & 0.5593 & \underline{0.6083} & 0.5501 & 0.5699  &  0.3329 & 0.5495 & 0.5219 & \textcolor{red!80!black}{\textbf{0.6167}} \\ 
&DOVER  & 0.7947 & 0.7789 & \underline{0.8248} & 0.7709 & 0.8185 & 0.5500 & 0.7851 & 0.8061 & \textcolor{red!80!black}{\textbf{0.8361}} \\

\hline
\multirow{6}{*}{\makecell[c]{YouHQ40}} 
&PSNR    & 23.99  & 22.67 & \underline{25.26}  & 25.01 & \textcolor{red!80!black}{\textbf{25.73}}  & 25.20  & 24.97   & 23.76  & 24.75 \\    
&SSIM    & 0.5999 & 0.5670 & 0.6857 & 0.6510 & \underline{0.6974} & 0.6865 & \textcolor{red!80!black}{\textbf{0.6984}}  & 0.6593 & 0.6954 \\ 
&MUSIQ   & 49.43 & 45.54 & 62.12 & 41.45 & \underline{62.27}  & 35.86 & 56.42   & 41.47  & \textcolor{red!80!black}{\textbf{62.71}} \\    
&CLIP-IQA& 0.3798 & 0.3442 & \underline{0.4661} & 0.3879 & 0.4520  & 0.2460  & 0.3960 & 0.3435 & \textcolor{red!80!black}{\textbf{0.4928}}\\
&MANIQA  & 0.5145 & 0.4710 & \underline{0.5539} & 0.4622 & 0.5500 & 0.3401 & 0.4717  & 0.4198 & \textcolor{red!80!black}{\textbf{0.5847}} \\ 
&DOVER  & 0.6914 & 0.6877 & \underline{0.8855} & 0.6347 & 0.8657 & 0.6216 & 0.8116  & 0.8084 & \textcolor{red!80!black}{\textbf{0.8872}} \\ 
\hline
\hline
\multirow{4}{*}{\makecell[c]{VideoLQ}} 
&MUSIQ  & 37.32  & 29.89 & \underline{62.48} & 37.94 & 58.66 & 30.88  & 55.99 & 54.59  & \textcolor{red!80!black}{\textbf{64.42}} \\    
&CLIP-IQA & 0.2766 & 0.2875 & \underline{0.3663} & 0.2763 & 0.3211  & 0.1817 & 0.3385 & 0.3126 & \textcolor{red!80!black}{\textbf{0.4599}}  \\
&MANIQA  & 0.4099 & 0.3447 & \underline{0.5721} & 0.4051 & 0.5477 & 0.3025  & 0.5004 & 0.5507 & \textcolor{red!80!black}{\textbf{0.5907}} \\ 
&DOVER & 0.5128 & 0.5141 & \underline{0.7645} & 0.5136 & 0.7399 & 0.4493 & 0.7254 & 0.7034 & \textcolor{red!80!black}{\textbf{0.7785}}  \\ 
\hline
\end{tabular}
\end{adjustbox}
\vspace{-0.6em}
\end{table*}

\begin{figure*}[t]
    \setlength{\abovecaptionskip}{0pt}
    \setlength{\belowcaptionskip}{-1pt}
    \centering
    \includegraphics[width=0.90\linewidth]{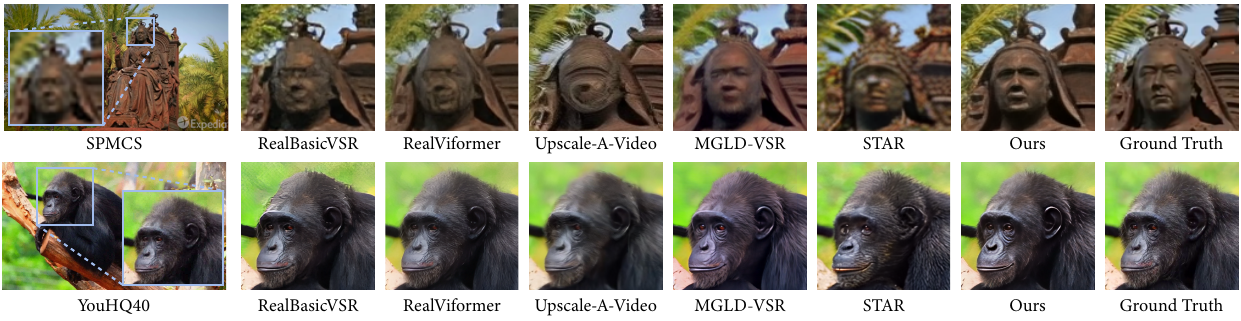}
    \caption{Visual comparisons on synthetic low-quality videos from SPMCS~\cite{yi2019progressive} and YouHQ40~\cite{zhou2024upscale} datasets.}
    \label{fig:compare}
    \vspace{-0.5em}
\end{figure*}

\section{Experiments}

\subsection{Experimental Settings}
\subsubsection{Datasets.} 
Similar to the most VSR approaches~\cite{chan2022investigating, yang2025motion, zhang2024realviformer}, 
we combine the training and validation sets from the REDS dataset~\cite{nah2019ntire} for model training, reserving four sequences (REDS4) for testing. 
We evaluate UltraVSR on four synthetic datasets (i.e., REDS4, SPMCS~\cite{yi2019progressive}, UDM10~\cite{tao2017detail}, and YouHQ40~\cite{zhou2024upscale}) and a real-world dataset (i.e., VideoLQ~\cite{chan2022investigating}) that contains 50 real-world video sequences with diverse degradations. 
For both synthetic training and testing datasets, LR-HR video pairs are generated following the degradation pipeline of RealBasicVSR~\cite{chan2022investigating}. 

\subsubsection{Training Details.} 
We implement UltraVSR using the PyTorch framework and the Diffusers~\cite{diffusers2022} library. 
The input sequence length is set to 6, with the spatial size cropped to 512 $\times$ 512 resolution. 
We freeze the original spatial modules of the pretrained Stable Diffusion model while only training the newly introduced LoRA and RTS modules. 
Adam~\cite{kingma2014adam} optimizer is employed as the optimizer with $\beta_1 = 0.9$ and $\beta_2 = 0.99$. 
We use a combination of MSE loss, perceptual loss, temporal GAN loss~\cite{zhou2023propainter}, and SJD regularization loss to train UltraVSR. 
The real and fake TRUs are trained by minimizing a standard denoising objective~\cite{ho2020denoising, vincent2011connection}. 
The weighting coefficient $\lambda$ is set to 0.5. 
The learning rate is set to $5 \times 10^{-5}$ for all models.

\begin{figure*}[t]
    \setlength{\abovecaptionskip}{0pt}
    \setlength{\belowcaptionskip}{-1pt}
    \centering
    \includegraphics[width=0.92\linewidth]{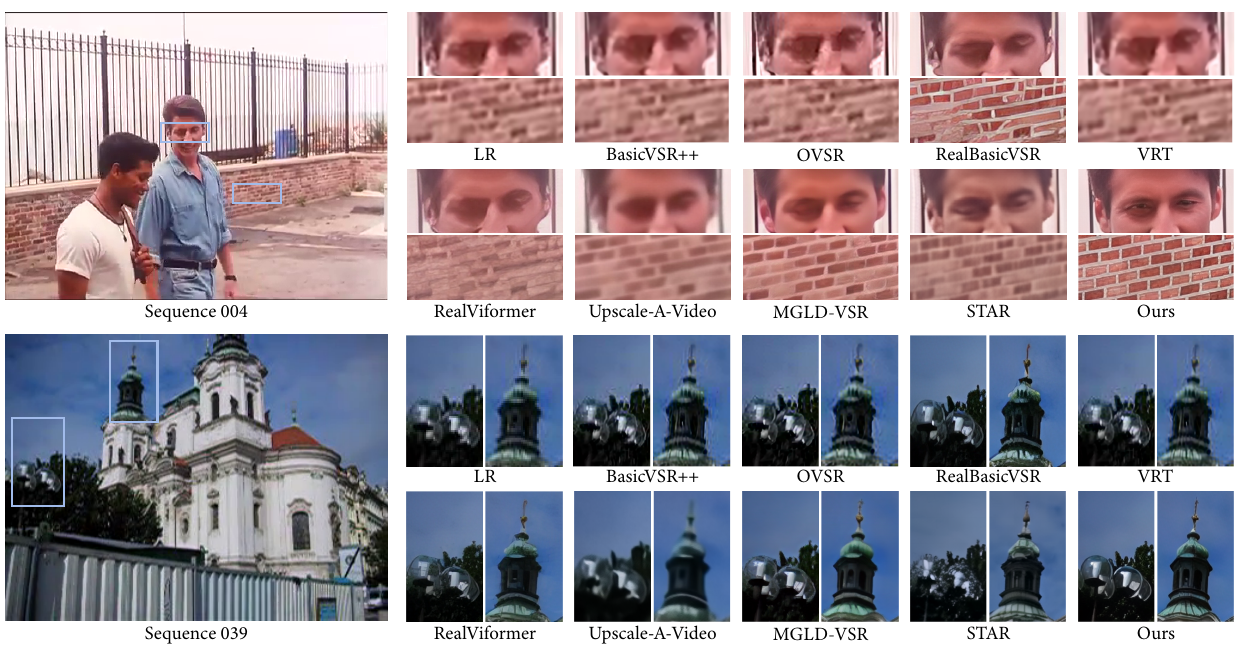}
    \caption{Visual comparisons on real-world low-quality videos from VideoLQ~\cite{chan2022investigating}dataset.}
    \label{fig:comparereal}
    \vspace{-0.5em}
\end{figure*}

\subsubsection{Evaluation Metrics.} 
We evaluate super-resolved video quality using three image-based quality metrics: MUSIQ~\cite{ke2021musiq}, CLIP-IQA~\cite{wang2023exploring}, and MANIQA~\cite{yang2022maniqa}, along with one video-specific quality metric, DOVER~\cite{wu2023exploring}. 
DOVER assesses both content quality and temporal consistency from both aesthetic and technical perspectives. 
In addition, the full-reference metrics (i.e., PSNR and SSIM) are also given as a reference. 
However, it should be noted that they poorly respond to realistic visuals~\cite{blau2018perception, gu2022ntire, ledig2017photo, yu2024scaling, jinjin2020pipal} and can not reliably evaluate the performance of diffusion-based VSR models. 

\subsection{Evaluation with State-of-the-art Methods}
To validate the efficacy of UltraVSR, we conduct both quantitative and qualitative comparisons with state-of-the-art VSR methods: 
BasicVSR++~\cite{chan2022basicvsr++}, 
OVSR~\cite{yi2021omniscient}, 
RealBasicVSR~\cite{chan2022investigating}, 
VRT~\cite{liang2024vrt}, 
RealViformer~\cite{zhang2024realviformer}, Upscale-A-Video~\cite{zhou2024upscale}, MGLD-VSR~\cite{shi2024motion}, and STAR~\cite{xie2025star}. 
Due to the extremely high memory requirements of STAR, we had to limit its output resolution to a maximum of 1440 pixels. 
In addition, we include comparisons with state-of-the-art SISR approaches (i.e., 
RealESRGAN~\cite{wang2021real}, 
StableSR~\cite{wang2024exploiting}, SUPIR~\cite{yu2024scaling}, ResShift~\cite{yue2024resshift} and OSEDiff~\cite{wu2024one}) in the supplementary materials.

\subsubsection{Quantitative Comparison.}
As demonstrated in Tab.~\ref{table:Quantitative}, we present the quantitative results with existing state-of-the-art VSR approaches. 
It can be seen that our UltraVSR consistently achieves the highest score across nearly all non-reference metrics on all synthetic benchmarks, demonstrating its superior performance. 
Although approaches like BasicVSR++~\cite{chan2022basicvsr++} and VRT~\cite{liang2024vrt} have better performance in terms of PSNR or SSIM, they usually produce unsatisfactory results on real-world degraded videos (see Sec.~\ref{qualitative}), as reflected in their lower no-reference metric scores. 
As for the real-world VSR dataset VideoLQ~\cite{chan2022investigating}, UltraVSR outperforms other methods across all non-reference metrics (MUSIQ: \textbf{+1.94}, CLIP-IQA: \textbf{+0.0936}, MANIQA: \textbf{+0.0186}, DOVER: \textbf{+0.014}). 
These notable improvements highlight UltraVSR’s strong ability to enhance complex and diverse real-world videos while preserving realistic details. 

\begin{table}[!t] 
\centering
\caption{Efficiency evaluations among diffusion-based VSR approaches. Both trainable/total parameters are measured for each model. The best results are \textcolor{red!80!black}{\textbf{highlighted}}. }
\vspace{-1em}
\label{table:compare}
\begin{adjustbox}{width=0.92\linewidth}
\begin{tabular}{c|cccc}
\hline
\multirow{2}{*}{\makecell[c]{Methods}} &  \multirow{2}{*}{\makecell[c]{Sampling step}} & \multicolumn{2}{c}{Runtime (s)}  & \# Params \\ 
\cline{3-4} 
& & 720$\times$1280 & 1440$\times$2560 & (M) \\
\hline
Upscale-A-Video~\cite{zhou2024upscale} & 30  & 16.1 & 32.0 & $\sim$ 96.6/746 \\    
MGLD-VSR~\cite{shi2024motion} & 50  & 31.8 & 204 & 156.0/1465 \\
STAR~\cite{xie2025star}  & 15  & 14.5 & 110.2 & 629.9/2139 \\
\hline
Ours &  \textcolor{red!80!black}{\textbf{1}} & \textcolor{red!80!black}{\textbf{0.89}} & \textcolor{red!80!black}{\textbf{2.67}} & \textcolor{red!80!black}{\textbf{10.5}}/1912  \\
\hline
\end{tabular}
\end{adjustbox}
\vspace{-1.0em}
\end{table}

\subsubsection{Qualitative Comparison.}
\label{qualitative}
To further assess the effectiveness of the proposed UltraVSR, we present qualitative comparisons on both synthetic datasets (SPMCS~\cite{yi2019progressive} and YouHQ40~\cite{zhou2024upscale} in Fig.~\ref{fig:compare}) and real-world dataset (VideoLQ~\cite{chan2022investigating} in Fig.~\ref{fig:comparereal}). 
On synthetic videos, it can be observed that the competing approaches struggle to reconstruct accurate texture (e.g., the results of Upscale-A-Video~\cite{zhou2024upscale} and STAR~\cite{xie2025star} in the first row) and blurriness (e.g., the results of RealViformer~\cite{zhang2024realviformer} and the result of Upscale-A-Video~\cite{zhou2024upscale} in the second row). 
In addition, MGLD-VSR~\cite{shi2024motion} introduces color artifacts in the reconstructed frames.  
In contrast, our UltraVSR excels in restoring lost textures and structures under complex degradation scenarios, such as the outlines of statues and animal fur. 
On real-world videos, UltraVSR demonstrates superior texture reconstruction capabilities compared to competing approaches. 
For instance, in sequence 004, it not only recovers clear and realistic facial expressions but also faithfully reconstructs the structural details of the surrounding brick wall. 
In sequence 039, UltraVSR successfully restores intricate architectural details and decorations, demonstrating its robustness and generalization ability under challenging degradation scenarios. 
More comparisons can be found in the supplementary materials. 

\begin{table*}[!t] 
\centering
\caption{Ablation study of different configurations. Both trainable/total parameters of the one-step reconstruction model are measured for each model variant. The best and second best results are \textcolor{red!80!black}{\textbf{highlighted}} and \underline{underlined}, respectively. }
\vspace{-0.8em}
\label{table:Quantitative2}
\begin{adjustbox}{width=0.90\linewidth}
\begin{tabular}{c|cccccc|cccc|c}
\hline
Models & DRS & RTS-Convolution Unit & RTS-Attention Unit& VSD~\cite{yin2024one, wang2023prolificdreamer} & SJD & TAI & MUSIQ & CLIP-IQA & MANIQA  & DOVER & \# Params (M) \\ 
\hline
A &  &  & & \usym{1F5F8} &  & \usym{1F5F8} & 0.5707  & 0.4360 & 61.93  & 0.7571 & 4.4/1804  \\    
B & \usym{1F5F8} &  &  & \usym{1F5F8} & & \usym{1F5F8} & 0.5759 & 0.4467 & 63.86 & 0.7564 & 4.4/1906 \\
C & \usym{1F5F8} & \usym{1F5F8} &  & \usym{1F5F8} & & \usym{1F5F8} & 0.5792 & 0.4547 & 63.93 & 0.7676 & 7.3/1909 \\
D & \usym{1F5F8} & \usym{1F5F8} & \usym{1F5F8} & \usym{1F5F8} &  &\usym{1F5F8} & 0.5824 & \underline{0.4572} & \underline{64.25} & \underline{0.7704} & 10.5/1912 \\
E & \usym{1F5F8} & \usym{1F5F8} & \usym{1F5F8} &  & \usym{1F5F8} &  & \underline{0.5851} & 0.4524 & 64.09 & 0.7665 & 10.5/1912 \\
\hline
Ours & \usym{1F5F8} & \usym{1F5F8} & \usym{1F5F8} & & \usym{1F5F8} & \usym{1F5F8} & \textcolor{red!80!black}{\textbf{0.5907}} & \textcolor{red!80!black}{\textbf{0.4599}} & \textcolor{red!80!black}{\textbf{64.42}} & \textcolor{red!80!black}{\textbf{0.7785}} & 10.5/1912 \\
\hline
\end{tabular}
\end{adjustbox}
\vspace{-1.5mm}
\end{table*}

\begin{figure*}[t]
    \setlength{\abovecaptionskip}{0pt}
    \setlength{\belowcaptionskip}{0pt}
    \centering
    \includegraphics[width=0.90\linewidth]{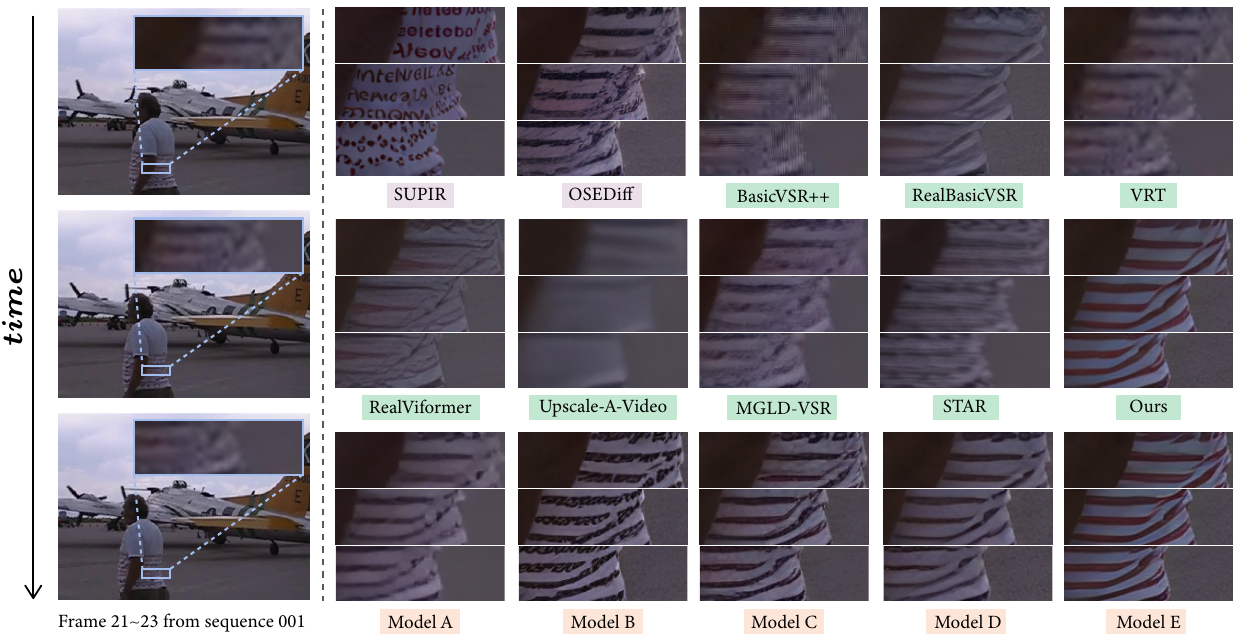}
    \caption{Temporal consistency analysis across \colorbox{sisrcolor}{SISR}, \colorbox{vsrcolor}{VSR} approaches, and different \colorbox{modelcolor}{model variants}.}
    \label{fig:consistency}
    \vspace{-0.8em}
\end{figure*}

\subsubsection{Temporal Consistency.}
%
From a quantitative perspective, we compare the DOVER metric, an effective video quality assessment approach, against competing methods in Tab.~\ref{table:Quantitative}. 
Due to the inherent randomness in the diffusion model, the DOVER scores of existing diffusion-based VSR methods are generally low. 
Benefiting from the proposed one-step diffusion space, our UltraVSR achieves the best score on all the datasets (e.g., \textbf{+0.011} on the SPMCS dataset and \textbf{+0.014} on the VideoLQ dataset). 
In addition, Fig.~\ref{fig:consistency} presents visual comparisons against both SISR and VSR approaches. 
It can be found that due to the inherent randomness and insufficient temporal modeling, results from SUPIR~\cite{yu2024scaling} and OSEDiff~\cite{wu2024one} exhibit severe temporal inconsistencies. 
While existing VSR approaches alleviate this issue, they often sacrifice texture details. 
In contrast, our ultraVSR successfully reconstructs videos with both fine texture structure and temporal coherence.

\subsubsection{Efficiency Evaluations.}
To evaluate the efficiency of UltraVSR, we compare its sampling step, runtime, and parameters against other diffusion-based VSR approaches in Tab.~\ref{table:compare}. 
we assess computational efficiency at two common resolutions: 720p (720$\times$1280) and 2K (1440$\times$2560). 
All models were evaluated on a single NVIDIA RTX 8000 GPU. 
Existing diffusion-based VSR approaches typically require dozens of iterative sampling steps, resulting in prohibitively long inference times, especially for 2K video, making them impractical for real-world deployment. 
Thanks to the efficient one-step diffusion space, our UltraVSR significantly reduces inference time, taking only \textbf{0.06$\times$} and \textbf{0.08$\times$} that of the second fastest approach at two resolutions. 
Moreover, UltraVSR has substantially fewer trainable parameters (just \textbf{10.5M}) compared to other diffusion-based methods, underscoring its efficiency during the training process.

\subsection{Ablation Study}
We conduct comprehensive experiments to evaluate the contributions of the key components in UltraVSR. 
The results are presented in Tab.~\ref{table:Quantitative2} and Fig.~\ref{fig:consistency}. 
We begin with Model A, which serves as our baseline, constructed by: 
(1) fixing the diffusion time step to 999, following the one-step generation paradigm established in prior works~\cite{yin2024one, nguyen2024swiftbrush, frans2024one, xie2024addsr}; 
(2) enabling only the trainable LoRA layers; and
(3) replacing our proposed Spatio-temporal Joint Distillation (SJD) with the conventional Variational Score Distillation (VSD)~\cite{yin2024one, wang2023prolificdreamer}. 
Due to its inability to dynamically perceive degradation levels, Model A exhibits limited reconstruction quality compared to Model B (with DRS). 
We incrementally enhance Model B with trainable RTS-convolution and RTS-attention units to form Models C and D, respectively, leading to steadily improved inter-frame consistency. 
We further evaluate the effectiveness of SJD and TAI by comparing Model D, Model E (with a fixed local sequence of length 10), and our model. 
It can be seen that SJD significantly enhances temporal coherence by introducing inter-frame difference supervision, leading to improved motion continuity across frames compared to VSD~\cite{yin2024one, wang2023prolificdreamer}. 
TAI further boosts performance by enabling long-range temporal modeling. 
%


\section{Conclusion}
We present UltraVSR, the first real-world VSR framework based on an efficient one-step diffusion space. 
It introduces Degradation-aware Reconstruction Scheduling (DRS) to transform multi-step denoising with one-step LR-to-HR reconstruction. 
To maintain temporal consistency, we develop the Recurrent Temporal Shift (RTS) module and Spatio-temporal Joint Distillation (SJD) for realistic and coherent outputs, along with Temporally Asynchronous Inference (TAI) for efficient long-range temporal modeling. 
Extensive experiments show that UltraVSR delivers favorable performance with significantly accelerated inference speed.

\section{Acknowledgments}
This work was supported in part by the National Major Science and Technology Projects of China under Grant 2009XJTU0016, in part by the National Natural Science Foundation of China under Grant U22B2049. 

\bibliographystyle{ACM-Reference-Format}
\bibliography{sample-base}


\begin{thebibliography}{65}


\ifx \showCODEN    \undefined \def \showCODEN     #1{\unskip}     \fi
\ifx \showISBNx    \undefined \def \showISBNx     #1{\unskip}     \fi
\ifx \showISBNxiii \undefined \def \showISBNxiii  #1{\unskip}     \fi
\ifx \showISSN     \undefined \def \showISSN      #1{\unskip}     \fi
\ifx \showLCCN     \undefined \def \showLCCN      #1{\unskip}     \fi
\ifx \shownote     \undefined \def \shownote      #1{#1}          \fi
\ifx \showarticletitle \undefined \def \showarticletitle #1{#1}   \fi
\ifx \showURL      \undefined \def \showURL       {\relax}        \fi
\providecommand\bibfield[2]{#2}
\providecommand\bibinfo[2]{#2}
\providecommand\natexlab[1]{#1}
\providecommand\showeprint[2][]{arXiv:#2}

\bibitem[Ai et~al\mbox{.}(2024)]%
        {ai2024skipvsr}
\bibfield{author}{\bibinfo{person}{Zekun Ai}, \bibinfo{person}{Xiaotong Luo},
  \bibinfo{person}{Yanyun Qu}, {and} \bibinfo{person}{Yuan Xie}.}
  \bibinfo{year}{2024}\natexlab{}.
\newblock \showarticletitle{SkipVSR: Adaptive Patch Routing for Video
  Super-Resolution with Inter-Frame Mask}. In
  \bibinfo{booktitle}{\emph{Proceedings of the 32nd ACM International
  Conference on Multimedia}}. \bibinfo{pages}{5874--5882}.
\newblock


\bibitem[Alaluf et~al\mbox{.}(2024)]%
        {alaluf2024cross}
\bibfield{author}{\bibinfo{person}{Yuval Alaluf}, \bibinfo{person}{Daniel
  Garibi}, \bibinfo{person}{Or Patashnik}, \bibinfo{person}{Hadar
  Averbuch-Elor}, {and} \bibinfo{person}{Daniel Cohen-Or}.}
  \bibinfo{year}{2024}\natexlab{}.
\newblock \showarticletitle{Cross-image attention for zero-shot appearance
  transfer}. In \bibinfo{booktitle}{\emph{ACM SIGGRAPH 2024 Conference
  Papers}}. \bibinfo{pages}{1--12}.
\newblock


\bibitem[Blattmann et~al\mbox{.}(2023)]%
        {blattmann2023align}
\bibfield{author}{\bibinfo{person}{Andreas Blattmann}, \bibinfo{person}{Robin
  Rombach}, \bibinfo{person}{Huan Ling}, \bibinfo{person}{Tim Dockhorn},
  \bibinfo{person}{Seung~Wook Kim}, \bibinfo{person}{Sanja Fidler}, {and}
  \bibinfo{person}{Karsten Kreis}.} \bibinfo{year}{2023}\natexlab{}.
\newblock \showarticletitle{Align your latents: High-resolution video synthesis
  with latent diffusion models}. In \bibinfo{booktitle}{\emph{Proceedings of
  the IEEE/CVF Conference on Computer Vision and Pattern Recognition}}.
  \bibinfo{pages}{22563--22575}.
\newblock


\bibitem[Blau and Michaeli(2018)]%
        {blau2018perception}
\bibfield{author}{\bibinfo{person}{Yochai Blau} {and} \bibinfo{person}{Tomer
  Michaeli}.} \bibinfo{year}{2018}\natexlab{}.
\newblock \showarticletitle{The perception-distortion tradeoff}. In
  \bibinfo{booktitle}{\emph{Proceedings of the IEEE conference on computer
  vision and pattern recognition}}. \bibinfo{pages}{6228--6237}.
\newblock


\bibitem[Cao et~al\mbox{.}(2023)]%
        {cao2023masactrl}
\bibfield{author}{\bibinfo{person}{Mingdeng Cao}, \bibinfo{person}{Xintao
  Wang}, \bibinfo{person}{Zhongang Qi}, \bibinfo{person}{Ying Shan},
  \bibinfo{person}{Xiaohu Qie}, {and} \bibinfo{person}{Yinqiang Zheng}.}
  \bibinfo{year}{2023}\natexlab{}.
\newblock \showarticletitle{Masactrl: Tuning-free mutual self-attention control
  for consistent image synthesis and editing}. In
  \bibinfo{booktitle}{\emph{Proceedings of the IEEE/CVF international
  conference on computer vision}}. \bibinfo{pages}{22560--22570}.
\newblock


\bibitem[Chan et~al\mbox{.}(2021)]%
        {chan2021basicvsr}
\bibfield{author}{\bibinfo{person}{Kelvin~CK Chan}, \bibinfo{person}{Xintao
  Wang}, \bibinfo{person}{Ke Yu}, \bibinfo{person}{Chao Dong}, {and}
  \bibinfo{person}{Chen~Change Loy}.} \bibinfo{year}{2021}\natexlab{}.
\newblock \showarticletitle{Basicvsr: The search for essential components in
  video super-resolution and beyond}. In \bibinfo{booktitle}{\emph{Proceedings
  of the IEEE/CVF conference on computer vision and pattern recognition}}.
  \bibinfo{pages}{4947--4956}.
\newblock


\bibitem[Chan et~al\mbox{.}(2022a)]%
        {chan2022basicvsr++}
\bibfield{author}{\bibinfo{person}{Kelvin~CK Chan}, \bibinfo{person}{Shangchen
  Zhou}, \bibinfo{person}{Xiangyu Xu}, {and} \bibinfo{person}{Chen~Change
  Loy}.} \bibinfo{year}{2022}\natexlab{a}.
\newblock \showarticletitle{Basicvsr++: Improving video super-resolution with
  enhanced propagation and alignment}. In \bibinfo{booktitle}{\emph{Proceedings
  of the IEEE/CVF conference on computer vision and pattern recognition}}.
  \bibinfo{pages}{5972--5981}.
\newblock


\bibitem[Chan et~al\mbox{.}(2022b)]%
        {chan2022investigating}
\bibfield{author}{\bibinfo{person}{Kelvin~CK Chan}, \bibinfo{person}{Shangchen
  Zhou}, \bibinfo{person}{Xiangyu Xu}, {and} \bibinfo{person}{Chen~Change
  Loy}.} \bibinfo{year}{2022}\natexlab{b}.
\newblock \showarticletitle{Investigating tradeoffs in real-world video
  super-resolution}. In \bibinfo{booktitle}{\emph{Proceedings of the IEEE/CVF
  Conference on Computer Vision and Pattern Recognition}}.
  \bibinfo{pages}{5962--5971}.
\newblock


\bibitem[Dhariwal and Nichol(2021)]%
        {dhariwal2021diffusion}
\bibfield{author}{\bibinfo{person}{Prafulla Dhariwal} {and}
  \bibinfo{person}{Alexander Nichol}.} \bibinfo{year}{2021}\natexlab{}.
\newblock \showarticletitle{Diffusion models beat gans on image synthesis}.
\newblock \bibinfo{journal}{\emph{Advances in neural information processing
  systems}}  \bibinfo{volume}{34} (\bibinfo{year}{2021}),
  \bibinfo{pages}{8780--8794}.
\newblock


\bibitem[Frans et~al\mbox{.}(2024)]%
        {frans2024one}
\bibfield{author}{\bibinfo{person}{Kevin Frans}, \bibinfo{person}{Danijar
  Hafner}, \bibinfo{person}{Sergey Levine}, {and} \bibinfo{person}{Pieter
  Abbeel}.} \bibinfo{year}{2024}\natexlab{}.
\newblock \showarticletitle{One step diffusion via shortcut models}.
\newblock \bibinfo{journal}{\emph{arXiv preprint arXiv:2410.12557}}
  (\bibinfo{year}{2024}).
\newblock


\bibitem[Fuoli et~al\mbox{.}(2019)]%
        {fuoli2019efficient}
\bibfield{author}{\bibinfo{person}{Dario Fuoli}, \bibinfo{person}{Shuhang Gu},
  {and} \bibinfo{person}{Radu Timofte}.} \bibinfo{year}{2019}\natexlab{}.
\newblock \showarticletitle{Efficient video super-resolution through recurrent
  latent space propagation}. In \bibinfo{booktitle}{\emph{2019 IEEE/CVF
  International Conference on Computer Vision Workshop (ICCVW)}}. IEEE,
  \bibinfo{pages}{3476--3485}.
\newblock


\bibitem[Geyer et~al\mbox{.}(2023)]%
        {geyer2023tokenflow}
\bibfield{author}{\bibinfo{person}{Michal Geyer}, \bibinfo{person}{Omer
  Bar-Tal}, \bibinfo{person}{Shai Bagon}, {and} \bibinfo{person}{Tali Dekel}.}
  \bibinfo{year}{2023}\natexlab{}.
\newblock \showarticletitle{Tokenflow: Consistent diffusion features for
  consistent video editing}.
\newblock \bibinfo{journal}{\emph{arXiv preprint arXiv:2307.10373}}
  (\bibinfo{year}{2023}).
\newblock


\bibitem[Gu et~al\mbox{.}(2022)]%
        {gu2022ntire}
\bibfield{author}{\bibinfo{person}{Jinjin Gu}, \bibinfo{person}{Haoming Cai},
  \bibinfo{person}{Chao Dong}, \bibinfo{person}{Jimmy~S Ren},
  \bibinfo{person}{Radu Timofte}, \bibinfo{person}{Yuan Gong},
  \bibinfo{person}{Shanshan Lao}, \bibinfo{person}{Shuwei Shi},
  \bibinfo{person}{Jiahao Wang}, \bibinfo{person}{Sidi Yang}, {et~al\mbox{.}}}
  \bibinfo{year}{2022}\natexlab{}.
\newblock \showarticletitle{NTIRE 2022 challenge on perceptual image quality
  assessment}. In \bibinfo{booktitle}{\emph{Proceedings of the IEEE/CVF
  conference on computer vision and pattern recognition}}.
  \bibinfo{pages}{951--967}.
\newblock


\bibitem[Gupta et~al\mbox{.}(2021)]%
        {gupta2021ada}
\bibfield{author}{\bibinfo{person}{Akash Gupta}, \bibinfo{person}{Padmaja
  Jonnalagedda}, \bibinfo{person}{Bir Bhanu}, {and} \bibinfo{person}{Amit~K
  Roy-Chowdhury}.} \bibinfo{year}{2021}\natexlab{}.
\newblock \showarticletitle{Ada-vsr: Adaptive video super-resolution with
  meta-learning}. In \bibinfo{booktitle}{\emph{Proceedings of the 29th ACM
  international conference on multimedia}}. \bibinfo{pages}{327--336}.
\newblock


\bibitem[Ho et~al\mbox{.}(2020)]%
        {ho2020denoising}
\bibfield{author}{\bibinfo{person}{Jonathan Ho}, \bibinfo{person}{Ajay Jain},
  {and} \bibinfo{person}{Pieter Abbeel}.} \bibinfo{year}{2020}\natexlab{}.
\newblock \showarticletitle{Denoising diffusion probabilistic models}.
\newblock \bibinfo{journal}{\emph{Advances in neural information processing
  systems}}  \bibinfo{volume}{33} (\bibinfo{year}{2020}),
  \bibinfo{pages}{6840--6851}.
\newblock


\bibitem[Hu et~al\mbox{.}(2022)]%
        {hu2022lora}
\bibfield{author}{\bibinfo{person}{Edward~J Hu}, \bibinfo{person}{Yelong Shen},
  \bibinfo{person}{Phillip Wallis}, \bibinfo{person}{Zeyuan Allen-Zhu},
  \bibinfo{person}{Yuanzhi Li}, \bibinfo{person}{Shean Wang},
  \bibinfo{person}{Lu Wang}, \bibinfo{person}{Weizhu Chen}, {et~al\mbox{.}}}
  \bibinfo{year}{2022}\natexlab{}.
\newblock \showarticletitle{Lora: Low-rank adaptation of large language
  models.}
\newblock \bibinfo{journal}{\emph{ICLR}} \bibinfo{volume}{1},
  \bibinfo{number}{2} (\bibinfo{year}{2022}), \bibinfo{pages}{3}.
\newblock


\bibitem[Jin et~al\mbox{.}(2023)]%
        {jin2023kernel}
\bibfield{author}{\bibinfo{person}{Shuo Jin}, \bibinfo{person}{Meiqin Liu},
  \bibinfo{person}{Chao Yao}, \bibinfo{person}{Chunyu Lin}, {and}
  \bibinfo{person}{Yao Zhao}.} \bibinfo{year}{2023}\natexlab{}.
\newblock \showarticletitle{Kernel dimension matters: To activate available
  kernels for real-time video super-resolution}. In
  \bibinfo{booktitle}{\emph{Proceedings of the 31st ACM International
  Conference on Multimedia}}. \bibinfo{pages}{8617--8625}.
\newblock


\bibitem[Jinjin et~al\mbox{.}(2020)]%
        {jinjin2020pipal}
\bibfield{author}{\bibinfo{person}{Gu Jinjin}, \bibinfo{person}{Cai Haoming},
  \bibinfo{person}{Chen Haoyu}, \bibinfo{person}{Ye Xiaoxing},
  \bibinfo{person}{Jimmy~S Ren}, {and} \bibinfo{person}{Dong Chao}.}
  \bibinfo{year}{2020}\natexlab{}.
\newblock \showarticletitle{Pipal: a large-scale image quality assessment
  dataset for perceptual image restoration}. In
  \bibinfo{booktitle}{\emph{Computer Vision--ECCV 2020: 16th European
  Conference, Glasgow, UK, August 23--28, 2020, Proceedings, Part XI 16}}.
  Springer, \bibinfo{pages}{633--651}.
\newblock


\bibitem[Kawar et~al\mbox{.}(2023)]%
        {kawar2023imagic}
\bibfield{author}{\bibinfo{person}{Bahjat Kawar}, \bibinfo{person}{Shiran
  Zada}, \bibinfo{person}{Oran Lang}, \bibinfo{person}{Omer Tov},
  \bibinfo{person}{Huiwen Chang}, \bibinfo{person}{Tali Dekel},
  \bibinfo{person}{Inbar Mosseri}, {and} \bibinfo{person}{Michal Irani}.}
  \bibinfo{year}{2023}\natexlab{}.
\newblock \showarticletitle{Imagic: Text-based real image editing with
  diffusion models}. In \bibinfo{booktitle}{\emph{Proceedings of the IEEE/CVF
  conference on computer vision and pattern recognition}}.
  \bibinfo{pages}{6007--6017}.
\newblock


\bibitem[Ke et~al\mbox{.}(2021)]%
        {ke2021musiq}
\bibfield{author}{\bibinfo{person}{Junjie Ke}, \bibinfo{person}{Qifei Wang},
  \bibinfo{person}{Yilin Wang}, \bibinfo{person}{Peyman Milanfar}, {and}
  \bibinfo{person}{Feng Yang}.} \bibinfo{year}{2021}\natexlab{}.
\newblock \showarticletitle{Musiq: Multi-scale image quality transformer}. In
  \bibinfo{booktitle}{\emph{Proceedings of the IEEE/CVF international
  conference on computer vision}}. \bibinfo{pages}{5148--5157}.
\newblock


\bibitem[Kingma and Ba(2014)]%
        {kingma2014adam}
\bibfield{author}{\bibinfo{person}{Diederik~P Kingma} {and}
  \bibinfo{person}{Jimmy Ba}.} \bibinfo{year}{2014}\natexlab{}.
\newblock \showarticletitle{Adam: A method for stochastic optimization}.
\newblock \bibinfo{journal}{\emph{arXiv preprint arXiv:1412.6980}}
  (\bibinfo{year}{2014}).
\newblock


\bibitem[Ledig et~al\mbox{.}(2017)]%
        {ledig2017photo}
\bibfield{author}{\bibinfo{person}{Christian Ledig}, \bibinfo{person}{Lucas
  Theis}, \bibinfo{person}{Ferenc Husz{\'a}r}, \bibinfo{person}{Jose
  Caballero}, \bibinfo{person}{Andrew Cunningham}, \bibinfo{person}{Alejandro
  Acosta}, \bibinfo{person}{Andrew Aitken}, \bibinfo{person}{Alykhan Tejani},
  \bibinfo{person}{Johannes Totz}, \bibinfo{person}{Zehan Wang},
  {et~al\mbox{.}}} \bibinfo{year}{2017}\natexlab{}.
\newblock \showarticletitle{Photo-realistic single image super-resolution using
  a generative adversarial network}. In \bibinfo{booktitle}{\emph{Proceedings
  of the IEEE conference on computer vision and pattern recognition}}.
  \bibinfo{pages}{4681--4690}.
\newblock


\bibitem[Leng et~al\mbox{.}(2022)]%
        {leng2022icnet}
\bibfield{author}{\bibinfo{person}{Jiaxu Leng}, \bibinfo{person}{Jia Wang},
  \bibinfo{person}{Xinbo Gao}, \bibinfo{person}{Bo Hu}, \bibinfo{person}{Ji
  Gan}, {and} \bibinfo{person}{Chenqiang Gao}.}
  \bibinfo{year}{2022}\natexlab{}.
\newblock \showarticletitle{Icnet: Joint alignment and reconstruction via
  iterative collaboration for video super-resolution}. In
  \bibinfo{booktitle}{\emph{Proceedings of the 30th ACM International
  Conference on Multimedia}}. \bibinfo{pages}{6675--6684}.
\newblock


\bibitem[Li et~al\mbox{.}(2023)]%
        {li2023simple}
\bibfield{author}{\bibinfo{person}{Dasong Li}, \bibinfo{person}{Xiaoyu Shi},
  \bibinfo{person}{Yi Zhang}, \bibinfo{person}{Ka~Chun Cheung},
  \bibinfo{person}{Simon See}, \bibinfo{person}{Xiaogang Wang},
  \bibinfo{person}{Hongwei Qin}, {and} \bibinfo{person}{Hongsheng Li}.}
  \bibinfo{year}{2023}\natexlab{}.
\newblock \showarticletitle{A simple baseline for video restoration with
  grouped spatial-temporal shift}. In \bibinfo{booktitle}{\emph{Proceedings of
  the IEEE/CVF Conference on Computer Vision and Pattern Recognition}}.
  \bibinfo{pages}{9822--9832}.
\newblock


\bibitem[Li et~al\mbox{.}(2025)]%
        {li2025generative}
\bibfield{author}{\bibinfo{person}{Yinchuan Li}, \bibinfo{person}{Xinyu Shao},
  \bibinfo{person}{Jianping Zhang}, \bibinfo{person}{Haozhi Wang},
  \bibinfo{person}{Leo~Maxime Brunswic}, \bibinfo{person}{Kaiwen Zhou},
  \bibinfo{person}{Jiqian Dong}, \bibinfo{person}{Kaiyang Guo},
  \bibinfo{person}{Xiu Li}, \bibinfo{person}{Zhitang Chen}, {et~al\mbox{.}}}
  \bibinfo{year}{2025}\natexlab{}.
\newblock \showarticletitle{Generative models in decision making: A survey}.
\newblock \bibinfo{journal}{\emph{arXiv preprint arXiv:2502.17100}}
  (\bibinfo{year}{2025}).
\newblock


\bibitem[Liang et~al\mbox{.}(2024)]%
        {liang2024vrt}
\bibfield{author}{\bibinfo{person}{Jingyun Liang}, \bibinfo{person}{Jiezhang
  Cao}, \bibinfo{person}{Yuchen Fan}, \bibinfo{person}{Kai Zhang},
  \bibinfo{person}{Rakesh Ranjan}, \bibinfo{person}{Yawei Li},
  \bibinfo{person}{Radu Timofte}, {and} \bibinfo{person}{Luc Van~Gool}.}
  \bibinfo{year}{2024}\natexlab{}.
\newblock \showarticletitle{Vrt: A video restoration transformer}.
\newblock \bibinfo{journal}{\emph{IEEE Transactions on Image Processing}}
  (\bibinfo{year}{2024}).
\newblock


\bibitem[Liang et~al\mbox{.}(2022)]%
        {liang2022recurrent}
\bibfield{author}{\bibinfo{person}{Jingyun Liang}, \bibinfo{person}{Yuchen
  Fan}, \bibinfo{person}{Xiaoyu Xiang}, \bibinfo{person}{Rakesh Ranjan},
  \bibinfo{person}{Eddy Ilg}, \bibinfo{person}{Simon Green},
  \bibinfo{person}{Jiezhang Cao}, \bibinfo{person}{Kai Zhang},
  \bibinfo{person}{Radu Timofte}, {and} \bibinfo{person}{Luc~V Gool}.}
  \bibinfo{year}{2022}\natexlab{}.
\newblock \showarticletitle{Recurrent video restoration transformer with guided
  deformable attention}.
\newblock \bibinfo{journal}{\emph{Advances in Neural Information Processing
  Systems}}  \bibinfo{volume}{35} (\bibinfo{year}{2022}),
  \bibinfo{pages}{378--393}.
\newblock


\bibitem[Lin et~al\mbox{.}(2019)]%
        {lin2019tsm}
\bibfield{author}{\bibinfo{person}{Ji Lin}, \bibinfo{person}{Chuang Gan}, {and}
  \bibinfo{person}{Song Han}.} \bibinfo{year}{2019}\natexlab{}.
\newblock \showarticletitle{Tsm: Temporal shift module for efficient video
  understanding}. In \bibinfo{booktitle}{\emph{Proceedings of the IEEE/CVF
  international conference on computer vision}}. \bibinfo{pages}{7083--7093}.
\newblock


\bibitem[Munoz et~al\mbox{.}(2021)]%
        {munoz2021temporal}
\bibfield{author}{\bibinfo{person}{Andres Munoz}, \bibinfo{person}{Mohammadreza
  Zolfaghari}, \bibinfo{person}{Max Argus}, {and} \bibinfo{person}{Thomas
  Brox}.} \bibinfo{year}{2021}\natexlab{}.
\newblock \showarticletitle{Temporal shift GAN for large scale video
  generation}. In \bibinfo{booktitle}{\emph{Proceedings of the IEEE/CVF Winter
  Conference on Applications of Computer Vision}}. \bibinfo{pages}{3179--3188}.
\newblock


\bibitem[Nah et~al\mbox{.}(2019)]%
        {nah2019ntire}
\bibfield{author}{\bibinfo{person}{Seungjun Nah}, \bibinfo{person}{Sungyong
  Baik}, \bibinfo{person}{Seokil Hong}, \bibinfo{person}{Gyeongsik Moon},
  \bibinfo{person}{Sanghyun Son}, \bibinfo{person}{Radu Timofte}, {and}
  \bibinfo{person}{Kyoung Mu~Lee}.} \bibinfo{year}{2019}\natexlab{}.
\newblock \showarticletitle{Ntire 2019 challenge on video deblurring and
  super-resolution: Dataset and study}. In
  \bibinfo{booktitle}{\emph{Proceedings of the IEEE/CVF conference on computer
  vision and pattern recognition workshops}}. \bibinfo{pages}{0--0}.
\newblock


\bibitem[Nguyen and Tran(2024)]%
        {nguyen2024swiftbrush}
\bibfield{author}{\bibinfo{person}{Thuan~Hoang Nguyen} {and}
  \bibinfo{person}{Anh Tran}.} \bibinfo{year}{2024}\natexlab{}.
\newblock \showarticletitle{Swiftbrush: One-step text-to-image diffusion model
  with variational score distillation}. In
  \bibinfo{booktitle}{\emph{Proceedings of the IEEE/CVF Conference on Computer
  Vision and Pattern Recognition}}. \bibinfo{pages}{7807--7816}.
\newblock


\bibitem[Patrick~von Platen and Wolf(2022)]%
        {diffusers2022}
\bibfield{author}{\bibinfo{person}{Anton Lozhkov Pedro Cuenca Nathan Lambert
  Kashif Rasul Mishig~Davaadorj Patrick~von Platen, Suraj~Patil} {and}
  \bibinfo{person}{Thomas Wolf}.} \bibinfo{year}{2022}\natexlab{}.
\newblock \showarticletitle{Diffusers: State-of-the-art Diffusion Models}.
\newblock


\bibitem[Poole et~al\mbox{.}(2022)]%
        {poole2022dreamfusion}
\bibfield{author}{\bibinfo{person}{Ben Poole}, \bibinfo{person}{Ajay Jain},
  \bibinfo{person}{Jonathan~T Barron}, {and} \bibinfo{person}{Ben Mildenhall}.}
  \bibinfo{year}{2022}\natexlab{}.
\newblock \showarticletitle{Dreamfusion: Text-to-3d using 2d diffusion}.
\newblock \bibinfo{journal}{\emph{arXiv preprint arXiv:2209.14988}}
  (\bibinfo{year}{2022}).
\newblock


\bibitem[Rombach et~al\mbox{.}(2022)]%
        {rombach2022high}
\bibfield{author}{\bibinfo{person}{Robin Rombach}, \bibinfo{person}{Andreas
  Blattmann}, \bibinfo{person}{Dominik Lorenz}, \bibinfo{person}{Patrick
  Esser}, {and} \bibinfo{person}{Bj{\"o}rn Ommer}.}
  \bibinfo{year}{2022}\natexlab{}.
\newblock \showarticletitle{High-resolution image synthesis with latent
  diffusion models}. In \bibinfo{booktitle}{\emph{Proceedings of the IEEE/CVF
  conference on computer vision and pattern recognition}}.
  \bibinfo{pages}{10684--10695}.
\newblock


\bibitem[Shi et~al\mbox{.}(2024)]%
        {shi2024motion}
\bibfield{author}{\bibinfo{person}{Xiaoyu Shi}, \bibinfo{person}{Zhaoyang
  Huang}, \bibinfo{person}{Fu-Yun Wang}, \bibinfo{person}{Weikang Bian},
  \bibinfo{person}{Dasong Li}, \bibinfo{person}{Yi Zhang},
  \bibinfo{person}{Manyuan Zhang}, \bibinfo{person}{Ka~Chun Cheung},
  \bibinfo{person}{Simon See}, \bibinfo{person}{Hongwei Qin}, {et~al\mbox{.}}}
  \bibinfo{year}{2024}\natexlab{}.
\newblock \showarticletitle{Motion-i2v: Consistent and controllable
  image-to-video generation with explicit motion modeling}. In
  \bibinfo{booktitle}{\emph{ACM SIGGRAPH 2024 Conference Papers}}.
  \bibinfo{pages}{1--11}.
\newblock


\bibitem[Sohl-Dickstein et~al\mbox{.}(2015)]%
        {sohl2015deep}
\bibfield{author}{\bibinfo{person}{Jascha Sohl-Dickstein},
  \bibinfo{person}{Eric Weiss}, \bibinfo{person}{Niru Maheswaranathan}, {and}
  \bibinfo{person}{Surya Ganguli}.} \bibinfo{year}{2015}\natexlab{}.
\newblock \showarticletitle{Deep unsupervised learning using nonequilibrium
  thermodynamics}. In \bibinfo{booktitle}{\emph{International conference on
  machine learning}}. PMLR, \bibinfo{pages}{2256--2265}.
\newblock


\bibitem[Song et~al\mbox{.}(2023)]%
        {song2023consistency}
\bibfield{author}{\bibinfo{person}{Yang Song}, \bibinfo{person}{Prafulla
  Dhariwal}, \bibinfo{person}{Mark Chen}, {and} \bibinfo{person}{Ilya
  Sutskever}.} \bibinfo{year}{2023}\natexlab{}.
\newblock \showarticletitle{Consistency models}.
\newblock  (\bibinfo{year}{2023}).
\newblock


\bibitem[Tang et~al\mbox{.}(2023)]%
        {tang2023emergent}
\bibfield{author}{\bibinfo{person}{Luming Tang}, \bibinfo{person}{Menglin Jia},
  \bibinfo{person}{Qianqian Wang}, \bibinfo{person}{Cheng~Perng Phoo}, {and}
  \bibinfo{person}{Bharath Hariharan}.} \bibinfo{year}{2023}\natexlab{}.
\newblock \showarticletitle{Emergent correspondence from image diffusion}.
\newblock \bibinfo{journal}{\emph{Advances in Neural Information Processing
  Systems}}  \bibinfo{volume}{36} (\bibinfo{year}{2023}),
  \bibinfo{pages}{1363--1389}.
\newblock


\bibitem[Tao et~al\mbox{.}(2017)]%
        {tao2017detail}
\bibfield{author}{\bibinfo{person}{Xin Tao}, \bibinfo{person}{Hongyun Gao},
  \bibinfo{person}{Renjie Liao}, \bibinfo{person}{Jue Wang}, {and}
  \bibinfo{person}{Jiaya Jia}.} \bibinfo{year}{2017}\natexlab{}.
\newblock \showarticletitle{Detail-revealing deep video super-resolution}. In
  \bibinfo{booktitle}{\emph{Proceedings of the IEEE international conference on
  computer vision}}. \bibinfo{pages}{4472--4480}.
\newblock


\bibitem[Vincent(2011)]%
        {vincent2011connection}
\bibfield{author}{\bibinfo{person}{Pascal Vincent}.}
  \bibinfo{year}{2011}\natexlab{}.
\newblock \showarticletitle{A connection between score matching and denoising
  autoencoders}.
\newblock \bibinfo{journal}{\emph{Neural computation}} \bibinfo{volume}{23},
  \bibinfo{number}{7} (\bibinfo{year}{2011}), \bibinfo{pages}{1661--1674}.
\newblock


\bibitem[Wang et~al\mbox{.}(2023a)]%
        {wang2023exploring}
\bibfield{author}{\bibinfo{person}{Jianyi Wang}, \bibinfo{person}{Kelvin~CK
  Chan}, {and} \bibinfo{person}{Chen~Change Loy}.}
  \bibinfo{year}{2023}\natexlab{a}.
\newblock \showarticletitle{Exploring clip for assessing the look and feel of
  images}. In \bibinfo{booktitle}{\emph{Proceedings of the AAAI conference on
  artificial intelligence}}, Vol.~\bibinfo{volume}{37}.
  \bibinfo{pages}{2555--2563}.
\newblock


\bibitem[Wang et~al\mbox{.}(2024a)]%
        {wang2024hero}
\bibfield{author}{\bibinfo{person}{Jiangang Wang}, \bibinfo{person}{Qingnan
  Fan}, \bibinfo{person}{Qi Zhang}, \bibinfo{person}{Haigen Liu},
  \bibinfo{person}{Yuhang Yu}, \bibinfo{person}{Jinwei Chen}, {and}
  \bibinfo{person}{Wenqi Ren}.} \bibinfo{year}{2024}\natexlab{a}.
\newblock \showarticletitle{Hero-SR: One-Step Diffusion for Super-Resolution
  with Human Perception Priors}.
\newblock \bibinfo{journal}{\emph{arXiv preprint arXiv:2412.07152}}
  (\bibinfo{year}{2024}).
\newblock


\bibitem[Wang et~al\mbox{.}(2024c)]%
        {wang2024exploiting}
\bibfield{author}{\bibinfo{person}{Jianyi Wang}, \bibinfo{person}{Zongsheng
  Yue}, \bibinfo{person}{Shangchen Zhou}, \bibinfo{person}{Kelvin~CK Chan},
  {and} \bibinfo{person}{Chen~Change Loy}.} \bibinfo{year}{2024}\natexlab{c}.
\newblock \showarticletitle{Exploiting diffusion prior for real-world image
  super-resolution}.
\newblock \bibinfo{journal}{\emph{International Journal of Computer Vision}}
  (\bibinfo{year}{2024}), \bibinfo{pages}{1--21}.
\newblock


\bibitem[Wang et~al\mbox{.}(2021)]%
        {wang2021real}
\bibfield{author}{\bibinfo{person}{Xintao Wang}, \bibinfo{person}{Liangbin
  Xie}, \bibinfo{person}{Chao Dong}, {and} \bibinfo{person}{Ying Shan}.}
  \bibinfo{year}{2021}\natexlab{}.
\newblock \showarticletitle{Real-esrgan: Training real-world blind
  super-resolution with pure synthetic data}. In
  \bibinfo{booktitle}{\emph{Proceedings of the IEEE/CVF international
  conference on computer vision}}. \bibinfo{pages}{1905--1914}.
\newblock


\bibitem[Wang et~al\mbox{.}(2024b)]%
        {wang2024sinsr}
\bibfield{author}{\bibinfo{person}{Yufei Wang}, \bibinfo{person}{Wenhan Yang},
  \bibinfo{person}{Xinyuan Chen}, \bibinfo{person}{Yaohui Wang},
  \bibinfo{person}{Lanqing Guo}, \bibinfo{person}{Lap-Pui Chau},
  \bibinfo{person}{Ziwei Liu}, \bibinfo{person}{Yu Qiao},
  \bibinfo{person}{Alex~C Kot}, {and} \bibinfo{person}{Bihan Wen}.}
  \bibinfo{year}{2024}\natexlab{b}.
\newblock \showarticletitle{Sinsr: diffusion-based image super-resolution in a
  single step}. In \bibinfo{booktitle}{\emph{Proceedings of the IEEE/CVF
  conference on computer vision and pattern recognition}}.
  \bibinfo{pages}{25796--25805}.
\newblock


\bibitem[Wang et~al\mbox{.}(2023b)]%
        {wang2023prolificdreamer}
\bibfield{author}{\bibinfo{person}{Zhengyi Wang}, \bibinfo{person}{Cheng Lu},
  \bibinfo{person}{Yikai Wang}, \bibinfo{person}{Fan Bao},
  \bibinfo{person}{Chongxuan Li}, \bibinfo{person}{Hang Su}, {and}
  \bibinfo{person}{Jun Zhu}.} \bibinfo{year}{2023}\natexlab{b}.
\newblock \showarticletitle{Prolificdreamer: High-fidelity and diverse
  text-to-3d generation with variational score distillation}.
\newblock \bibinfo{journal}{\emph{Advances in Neural Information Processing
  Systems}}  \bibinfo{volume}{36} (\bibinfo{year}{2023}),
  \bibinfo{pages}{8406--8441}.
\newblock


\bibitem[Wu et~al\mbox{.}(2023)]%
        {wu2023exploring}
\bibfield{author}{\bibinfo{person}{Haoning Wu}, \bibinfo{person}{Erli Zhang},
  \bibinfo{person}{Liang Liao}, \bibinfo{person}{Chaofeng Chen},
  \bibinfo{person}{Jingwen Hou}, \bibinfo{person}{Annan Wang},
  \bibinfo{person}{Wenxiu Sun}, \bibinfo{person}{Qiong Yan}, {and}
  \bibinfo{person}{Weisi Lin}.} \bibinfo{year}{2023}\natexlab{}.
\newblock \showarticletitle{Exploring video quality assessment on user
  generated contents from aesthetic and technical perspectives}. In
  \bibinfo{booktitle}{\emph{Proceedings of the IEEE/CVF International
  Conference on Computer Vision}}. \bibinfo{pages}{20144--20154}.
\newblock


\bibitem[Wu et~al\mbox{.}(2024)]%
        {wu2024one}
\bibfield{author}{\bibinfo{person}{Rongyuan Wu}, \bibinfo{person}{Lingchen
  Sun}, \bibinfo{person}{Zhiyuan Ma}, {and} \bibinfo{person}{Lei Zhang}.}
  \bibinfo{year}{2024}\natexlab{}.
\newblock \showarticletitle{One-step effective diffusion network for real-world
  image super-resolution}.
\newblock \bibinfo{journal}{\emph{Advances in Neural Information Processing
  Systems}}  \bibinfo{volume}{37} (\bibinfo{year}{2024}),
  \bibinfo{pages}{92529--92553}.
\newblock


\bibitem[Xiao et~al\mbox{.}(2024)]%
        {xiao2024asymmetric}
\bibfield{author}{\bibinfo{person}{Zeyu Xiao}, \bibinfo{person}{Dachun Kai},
  \bibinfo{person}{Yueyi Zhang}, \bibinfo{person}{Xiaoyan Sun}, {and}
  \bibinfo{person}{Zhiwei Xiong}.} \bibinfo{year}{2024}\natexlab{}.
\newblock \showarticletitle{Asymmetric Event-Guided Video Super-Resolution}. In
  \bibinfo{booktitle}{\emph{Proceedings of the 32nd ACM International
  Conference on Multimedia}}. \bibinfo{pages}{2409--2418}.
\newblock


\bibitem[Xiao et~al\mbox{.}(2020)]%
        {xiao2020space}
\bibfield{author}{\bibinfo{person}{Zeyu Xiao}, \bibinfo{person}{Zhiwei Xiong},
  \bibinfo{person}{Xueyang Fu}, \bibinfo{person}{Dong Liu}, {and}
  \bibinfo{person}{Zheng-Jun Zha}.} \bibinfo{year}{2020}\natexlab{}.
\newblock \showarticletitle{Space-time video super-resolution using temporal
  profiles}. In \bibinfo{booktitle}{\emph{Proceedings of the 28th ACM
  International Conference on Multimedia}}. \bibinfo{pages}{664--672}.
\newblock


\bibitem[Xie et~al\mbox{.}(2025)]%
        {xie2025star}
\bibfield{author}{\bibinfo{person}{Rui Xie}, \bibinfo{person}{Yinhong Liu},
  \bibinfo{person}{Penghao Zhou}, \bibinfo{person}{Chen Zhao},
  \bibinfo{person}{Jun Zhou}, \bibinfo{person}{Kai Zhang},
  \bibinfo{person}{Zhenyu Zhang}, \bibinfo{person}{Jian Yang},
  \bibinfo{person}{Zhenheng Yang}, {and} \bibinfo{person}{Ying Tai}.}
  \bibinfo{year}{2025}\natexlab{}.
\newblock \showarticletitle{STAR: Spatial-Temporal Augmentation with
  Text-to-Video Models for Real-World Video Super-Resolution}.
\newblock \bibinfo{journal}{\emph{arXiv preprint arXiv:2501.02976}}
  (\bibinfo{year}{2025}).
\newblock


\bibitem[Xie et~al\mbox{.}(2024)]%
        {xie2024addsr}
\bibfield{author}{\bibinfo{person}{Rui Xie}, \bibinfo{person}{Chen Zhao},
  \bibinfo{person}{Kai Zhang}, \bibinfo{person}{Zhenyu Zhang},
  \bibinfo{person}{Jun Zhou}, \bibinfo{person}{Jian Yang}, {and}
  \bibinfo{person}{Ying Tai}.} \bibinfo{year}{2024}\natexlab{}.
\newblock \showarticletitle{Addsr: Accelerating diffusion-based blind
  super-resolution with adversarial diffusion distillation}.
\newblock \bibinfo{journal}{\emph{arXiv preprint arXiv:2404.01717}}
  (\bibinfo{year}{2024}).
\newblock


\bibitem[Yang et~al\mbox{.}(2023)]%
        {yang2023paint}
\bibfield{author}{\bibinfo{person}{Binxin Yang}, \bibinfo{person}{Shuyang Gu},
  \bibinfo{person}{Bo Zhang}, \bibinfo{person}{Ting Zhang},
  \bibinfo{person}{Xuejin Chen}, \bibinfo{person}{Xiaoyan Sun},
  \bibinfo{person}{Dong Chen}, {and} \bibinfo{person}{Fang Wen}.}
  \bibinfo{year}{2023}\natexlab{}.
\newblock \showarticletitle{Paint by example: Exemplar-based image editing with
  diffusion models}. In \bibinfo{booktitle}{\emph{Proceedings of the IEEE/CVF
  Conference on Computer Vision and Pattern Recognition}}.
  \bibinfo{pages}{18381--18391}.
\newblock


\bibitem[Yang et~al\mbox{.}(2022)]%
        {yang2022maniqa}
\bibfield{author}{\bibinfo{person}{Sidi Yang}, \bibinfo{person}{Tianhe Wu},
  \bibinfo{person}{Shuwei Shi}, \bibinfo{person}{Shanshan Lao},
  \bibinfo{person}{Yuan Gong}, \bibinfo{person}{Mingdeng Cao},
  \bibinfo{person}{Jiahao Wang}, {and} \bibinfo{person}{Yujiu Yang}.}
  \bibinfo{year}{2022}\natexlab{}.
\newblock \showarticletitle{Maniqa: Multi-dimension attention network for
  no-reference image quality assessment}. In
  \bibinfo{booktitle}{\emph{Proceedings of the IEEE/CVF conference on computer
  vision and pattern recognition}}. \bibinfo{pages}{1191--1200}.
\newblock


\bibitem[Yang et~al\mbox{.}(2025)]%
        {yang2025motion}
\bibfield{author}{\bibinfo{person}{Xi Yang}, \bibinfo{person}{Chenhang He},
  \bibinfo{person}{Jianqi Ma}, {and} \bibinfo{person}{Lei Zhang}.}
  \bibinfo{year}{2025}\natexlab{}.
\newblock \showarticletitle{Motion-guided latent diffusion for temporally
  consistent real-world video super-resolution}. In
  \bibinfo{booktitle}{\emph{European Conference on Computer Vision}}. Springer,
  \bibinfo{pages}{224--242}.
\newblock


\bibitem[Yang et~al\mbox{.}(2024)]%
        {yang2024cogvideox}
\bibfield{author}{\bibinfo{person}{Zhuoyi Yang}, \bibinfo{person}{Jiayan Teng},
  \bibinfo{person}{Wendi Zheng}, \bibinfo{person}{Ming Ding},
  \bibinfo{person}{Shiyu Huang}, \bibinfo{person}{Jiazheng Xu},
  \bibinfo{person}{Yuanming Yang}, \bibinfo{person}{Wenyi Hong},
  \bibinfo{person}{Xiaohan Zhang}, \bibinfo{person}{Guanyu Feng},
  {et~al\mbox{.}}} \bibinfo{year}{2024}\natexlab{}.
\newblock \showarticletitle{Cogvideox: Text-to-video diffusion models with an
  expert transformer}.
\newblock \bibinfo{journal}{\emph{arXiv preprint arXiv:2408.06072}}
  (\bibinfo{year}{2024}).
\newblock


\bibitem[Yi et~al\mbox{.}(2021)]%
        {yi2021omniscient}
\bibfield{author}{\bibinfo{person}{Peng Yi}, \bibinfo{person}{Zhongyuan Wang},
  \bibinfo{person}{Kui Jiang}, \bibinfo{person}{Junjun Jiang},
  \bibinfo{person}{Tao Lu}, \bibinfo{person}{Xin Tian}, {and}
  \bibinfo{person}{Jiayi Ma}.} \bibinfo{year}{2021}\natexlab{}.
\newblock \showarticletitle{Omniscient video super-resolution}. In
  \bibinfo{booktitle}{\emph{Proceedings of the IEEE/CVF international
  conference on computer vision}}. \bibinfo{pages}{4429--4438}.
\newblock


\bibitem[Yi et~al\mbox{.}(2019)]%
        {yi2019progressive}
\bibfield{author}{\bibinfo{person}{Peng Yi}, \bibinfo{person}{Zhongyuan Wang},
  \bibinfo{person}{Kui Jiang}, \bibinfo{person}{Junjun Jiang}, {and}
  \bibinfo{person}{Jiayi Ma}.} \bibinfo{year}{2019}\natexlab{}.
\newblock \showarticletitle{Progressive fusion video super-resolution network
  via exploiting non-local spatio-temporal correlations}. In
  \bibinfo{booktitle}{\emph{Proceedings of the IEEE/CVF international
  conference on computer vision}}. \bibinfo{pages}{3106--3115}.
\newblock


\bibitem[Yin et~al\mbox{.}(2024)]%
        {yin2024one}
\bibfield{author}{\bibinfo{person}{Tianwei Yin}, \bibinfo{person}{Micha{\"e}l
  Gharbi}, \bibinfo{person}{Richard Zhang}, \bibinfo{person}{Eli Shechtman},
  \bibinfo{person}{Fredo Durand}, \bibinfo{person}{William~T Freeman}, {and}
  \bibinfo{person}{Taesung Park}.} \bibinfo{year}{2024}\natexlab{}.
\newblock \showarticletitle{One-step diffusion with distribution matching
  distillation}. In \bibinfo{booktitle}{\emph{Proceedings of the IEEE/CVF
  conference on computer vision and pattern recognition}}.
  \bibinfo{pages}{6613--6623}.
\newblock


\bibitem[Yu et~al\mbox{.}(2024)]%
        {yu2024scaling}
\bibfield{author}{\bibinfo{person}{Fanghua Yu}, \bibinfo{person}{Jinjin Gu},
  \bibinfo{person}{Zheyuan Li}, \bibinfo{person}{Jinfan Hu},
  \bibinfo{person}{Xiangtao Kong}, \bibinfo{person}{Xintao Wang},
  \bibinfo{person}{Jingwen He}, \bibinfo{person}{Yu Qiao}, {and}
  \bibinfo{person}{Chao Dong}.} \bibinfo{year}{2024}\natexlab{}.
\newblock \showarticletitle{Scaling up to excellence: Practicing model scaling
  for photo-realistic image restoration in the wild}. In
  \bibinfo{booktitle}{\emph{Proceedings of the IEEE/CVF Conference on Computer
  Vision and Pattern Recognition}}. \bibinfo{pages}{25669--25680}.
\newblock


\bibitem[Yue et~al\mbox{.}(2024)]%
        {yue2024resshift}
\bibfield{author}{\bibinfo{person}{Zongsheng Yue}, \bibinfo{person}{Jianyi
  Wang}, {and} \bibinfo{person}{Chen~Change Loy}.}
  \bibinfo{year}{2024}\natexlab{}.
\newblock \showarticletitle{Resshift: Efficient diffusion model for image
  super-resolution by residual shifting}.
\newblock \bibinfo{journal}{\emph{Advances in Neural Information Processing
  Systems}}  \bibinfo{volume}{36} (\bibinfo{year}{2024}).
\newblock


\bibitem[Zhang et~al\mbox{.}(2023)]%
        {zhang2023adding}
\bibfield{author}{\bibinfo{person}{Lvmin Zhang}, \bibinfo{person}{Anyi Rao},
  {and} \bibinfo{person}{Maneesh Agrawala}.} \bibinfo{year}{2023}\natexlab{}.
\newblock \showarticletitle{Adding conditional control to text-to-image
  diffusion models}. In \bibinfo{booktitle}{\emph{Proceedings of the IEEE/CVF
  International Conference on Computer Vision}}. \bibinfo{pages}{3836--3847}.
\newblock


\bibitem[Zhang and Yao(2024)]%
        {zhang2024realviformer}
\bibfield{author}{\bibinfo{person}{Yuehan Zhang} {and} \bibinfo{person}{Angela
  Yao}.} \bibinfo{year}{2024}\natexlab{}.
\newblock \showarticletitle{Realviformer: Investigating attention for
  real-world video super-resolution}. In \bibinfo{booktitle}{\emph{European
  Conference on Computer Vision}}. Springer, \bibinfo{pages}{412--428}.
\newblock


\bibitem[Zhou et~al\mbox{.}(2023)]%
        {zhou2023propainter}
\bibfield{author}{\bibinfo{person}{Shangchen Zhou}, \bibinfo{person}{Chongyi
  Li}, \bibinfo{person}{Kelvin~CK Chan}, {and} \bibinfo{person}{Chen~Change
  Loy}.} \bibinfo{year}{2023}\natexlab{}.
\newblock \showarticletitle{Propainter: Improving propagation and transformer
  for video inpainting}. In \bibinfo{booktitle}{\emph{Proceedings of the
  IEEE/CVF international conference on computer vision}}.
  \bibinfo{pages}{10477--10486}.
\newblock


\bibitem[Zhou et~al\mbox{.}(2024)]%
        {zhou2024upscale}
\bibfield{author}{\bibinfo{person}{Shangchen Zhou}, \bibinfo{person}{Peiqing
  Yang}, \bibinfo{person}{Jianyi Wang}, \bibinfo{person}{Yihang Luo}, {and}
  \bibinfo{person}{Chen~Change Loy}.} \bibinfo{year}{2024}\natexlab{}.
\newblock \showarticletitle{Upscale-A-Video: Temporal-Consistent Diffusion
  Model for Real-World Video Super-Resolution}. In
  \bibinfo{booktitle}{\emph{Proceedings of the IEEE/CVF Conference on Computer
  Vision and Pattern Recognition}}. \bibinfo{pages}{2535--2545}.
\newblock


\end{thebibliography}



\begin{thebibliography}{6}


\ifx \showCODEN    \undefined \def \showCODEN     #1{\unskip}     \fi
\ifx \showISBNx    \undefined \def \showISBNx     #1{\unskip}     \fi
\ifx \showISBNxiii \undefined \def \showISBNxiii  #1{\unskip}     \fi
\ifx \showISSN     \undefined \def \showISSN      #1{\unskip}     \fi
\ifx \showLCCN     \undefined \def \showLCCN      #1{\unskip}     \fi
\ifx \shownote     \undefined \def \shownote      #1{#1}          \fi
\ifx \showarticletitle \undefined \def \showarticletitle #1{#1}   \fi
\ifx \showURL      \undefined \def \showURL       {\relax}        \fi
\providecommand\bibfield[2]{#2}
\providecommand\bibinfo[2]{#2}
\providecommand\natexlab[1]{#1}
\providecommand\showeprint[2][]{arXiv:#2}

\bibitem[Chan et~al\mbox{.}(2022)]%
        {chan2022investigating}
\bibfield{author}{\bibinfo{person}{Kelvin~CK Chan}, \bibinfo{person}{Shangchen
  Zhou}, \bibinfo{person}{Xiangyu Xu}, {and} \bibinfo{person}{Chen~Change
  Loy}.} \bibinfo{year}{2022}\natexlab{}.
\newblock \showarticletitle{Investigating tradeoffs in real-world video
  super-resolution}. In \bibinfo{booktitle}{\emph{Proceedings of the IEEE/CVF
  Conference on Computer Vision and Pattern Recognition}}.
  \bibinfo{pages}{5962--5971}.
\newblock


\bibitem[Hu et~al\mbox{.}(2022)]%
        {hu2022lora}
\bibfield{author}{\bibinfo{person}{Edward~J Hu}, \bibinfo{person}{Yelong Shen},
  \bibinfo{person}{Phillip Wallis}, \bibinfo{person}{Zeyuan Allen-Zhu},
  \bibinfo{person}{Yuanzhi Li}, \bibinfo{person}{Shean Wang},
  \bibinfo{person}{Lu Wang}, \bibinfo{person}{Weizhu Chen}, {et~al\mbox{.}}}
  \bibinfo{year}{2022}\natexlab{}.
\newblock \showarticletitle{Lora: Low-rank adaptation of large language
  models.}
\newblock \bibinfo{journal}{\emph{ICLR}} \bibinfo{volume}{1},
  \bibinfo{number}{2} (\bibinfo{year}{2022}), \bibinfo{pages}{3}.
\newblock


\bibitem[Rombach et~al\mbox{.}(2022)]%
        {rombach2022high}
\bibfield{author}{\bibinfo{person}{Robin Rombach}, \bibinfo{person}{Andreas
  Blattmann}, \bibinfo{person}{Dominik Lorenz}, \bibinfo{person}{Patrick
  Esser}, {and} \bibinfo{person}{Bj{\"o}rn Ommer}.}
  \bibinfo{year}{2022}\natexlab{}.
\newblock \showarticletitle{High-resolution image synthesis with latent
  diffusion models}. In \bibinfo{booktitle}{\emph{Proceedings of the IEEE/CVF
  conference on computer vision and pattern recognition}}.
  \bibinfo{pages}{10684--10695}.
\newblock


\bibitem[Xie et~al\mbox{.}(2025)]%
        {xie2025star}
\bibfield{author}{\bibinfo{person}{Rui Xie}, \bibinfo{person}{Yinhong Liu},
  \bibinfo{person}{Penghao Zhou}, \bibinfo{person}{Chen Zhao},
  \bibinfo{person}{Jun Zhou}, \bibinfo{person}{Kai Zhang},
  \bibinfo{person}{Zhenyu Zhang}, \bibinfo{person}{Jian Yang},
  \bibinfo{person}{Zhenheng Yang}, {and} \bibinfo{person}{Ying Tai}.}
  \bibinfo{year}{2025}\natexlab{}.
\newblock \showarticletitle{STAR: Spatial-Temporal Augmentation with
  Text-to-Video Models for Real-World Video Super-Resolution}.
\newblock \bibinfo{journal}{\emph{arXiv preprint arXiv:2501.02976}}
  (\bibinfo{year}{2025}).
\newblock


\bibitem[Yang et~al\mbox{.}(2025)]%
        {yang2025motion}
\bibfield{author}{\bibinfo{person}{Xi Yang}, \bibinfo{person}{Chenhang He},
  \bibinfo{person}{Jianqi Ma}, {and} \bibinfo{person}{Lei Zhang}.}
  \bibinfo{year}{2025}\natexlab{}.
\newblock \showarticletitle{Motion-guided latent diffusion for temporally
  consistent real-world video super-resolution}. In
  \bibinfo{booktitle}{\emph{European Conference on Computer Vision}}. Springer,
  \bibinfo{pages}{224--242}.
\newblock


\bibitem[Zhou et~al\mbox{.}(2024)]%
        {zhou2024upscale}
\bibfield{author}{\bibinfo{person}{Shangchen Zhou}, \bibinfo{person}{Peiqing
  Yang}, \bibinfo{person}{Jianyi Wang}, \bibinfo{person}{Yihang Luo}, {and}
  \bibinfo{person}{Chen~Change Loy}.} \bibinfo{year}{2024}\natexlab{}.
\newblock \showarticletitle{Upscale-A-Video: Temporal-Consistent Diffusion
  Model for Real-World Video Super-Resolution}. In
  \bibinfo{booktitle}{\emph{Proceedings of the IEEE/CVF Conference on Computer
  Vision and Pattern Recognition}}. \bibinfo{pages}{2535--2545}.
\newblock


\end{thebibliography}

\end{document}


\title[UltraVSR: Achieving Ultra-Realistic Video Super-Resolution with Efficient One-Step Diffusion Space]{Supplementary Material for "UltraVSR: Achieving Ultra-Realistic Video Super-Resolution with Efficient One-Step Diffusion Space"}

        \author{Yong Liu}
        \affiliation{
          \institution{Xi'an Jiaotong University}
          \city{Xi'an}
          \country{China}
        }
        \email{liuy1996v@qq.com}
        
        \author{Jinshan Pan}
        \affiliation{
          \institution{Nanjing University of Science and Technology}
          \city{Nanjing}
          \country{China}
        }
        \email{sdluran@gmail.com}
        
        \author{Yinchuan Li}
        \affiliation{
          \institution{Huawei Noah's Ark Lab}
          \city{Beijing}
          \country{China}
        }

        \author{Qingji Dong}
        \affiliation{
          \institution{Xi'an Jiaotong University}
          \city{Xi'an}
          \country{China}
        }

        \author{Chao Zhu}
        \affiliation{
          \institution{Xi'an Jiaotong University}
          \city{Xi'an}
          \country{China}
        }

        \author{Yu Guo}
        \affiliation{
          \institution{Xi'an Jiaotong University}
          \city{Xi'an}
          \country{China}
        }

        \author{Fei Wang}
        \authornote{Corresponding author. Author affiliation: National Key Laboratory of Human-Machine Hybrid Augmented Intelligence, National Engineering Research Center of Visual Information and Applications, Institute of Artificial Intelligence and Robotics, Xi'an Jiaotong University.}
        \affiliation{
          \institution{Xi'an Jiaotong University}
          \city{Xi'an}
          \country{China}
        }  
        \email{wfx@xjtu.edu.cn}

\ccsdesc[500]{Computing methodologies~Reconstruction}
\ccsdesc[300]{Computing methodologies~Image processing}
\ccsdesc[100]{Computing methodologies~Computational photography}

\keywords{Video super-resolution, Diffusion model, Temporal consistency}

\begin{teaserfigure}
  \centering
  \includegraphics[width=0.92\textwidth]{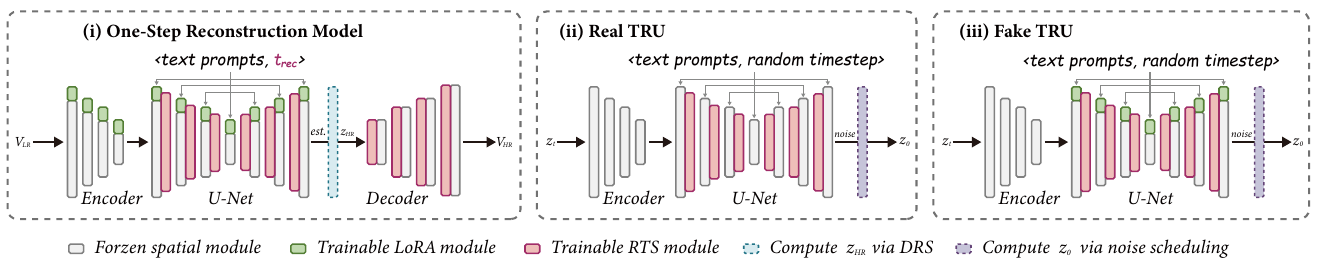}
  \vspace{-1.2em}
  \caption{The detailed network structure of different models in the proposed UltraVSR.}
  \label{fig:teaser}
\end{teaserfigure}

\maketitle

\section{Network Details}
To address the challenges of real-world VSR, we present UltraVSR, an efficient one-step diffusion-based framework.
UltraVSR comprises three core components: a one-step reconstruction model, a Real Temporal Regularized UNet (Real TRU), and a Fake Temporal Regularized UNet (Fake TRU). 
Their network structures are illustrated in Fig.~\ref{fig:teaser}.

\subsection{One-step Reconstruction Model}
By introducing a Degradation-aware Reconstruction Scheduling (DRS), our one-step reconstruction model enables direct one-step restoration from LR to HR videos. 
To adapt the pretrained generative prior to this paradigm, we integrate LoRA layers~\cite{hu2022lora} into both the VAE encoder and the diffusion UNet. 
Moreover, since the inconsistency details in Stable Diffusion primarily stem from the diffusion UNet and VAE decoder~\cite{yang2025motion,zhou2024upscale}, we develop a lightweight yet effective Recurrent Temporal Shift (RTS) module to capture temporal consistency across frames and insert it into each scale of both the diffusion UNet and VAE decoder. 

\begin{figure*}[t]
    \setlength{\abovecaptionskip}{0pt}
    \setlength{\belowcaptionskip}{-1pt}
    \centering
    \includegraphics[width=0.80\linewidth]{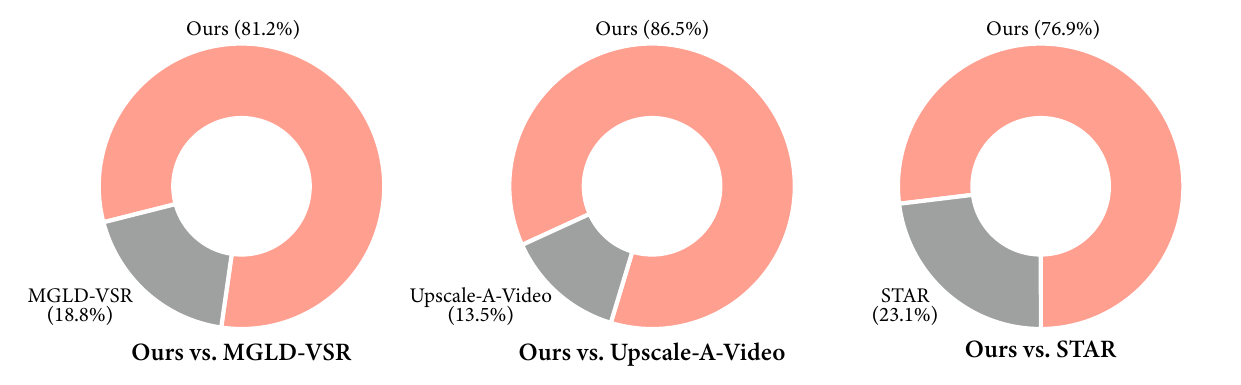}
    \caption{User study results. Our UltraVSR is consistently preferred by human evaluators over other methods.}
    \label{fig:user}
    \vspace{-0.8em}
\end{figure*}

\subsection{Real TRU and Fake TRU}
To guide the one-step reconstruction model toward generating visually realistic and temporally consistent videos, we propose Spatio-temporal Joint Distillation (SJD), comprising a Real TRU and a Fake TRU. 
Both TRUs are adapted from the pretrained Stable Diffusion model~\cite{rombach2022high} and updated to learn real/fake video distribution based on the samples from ground truth and reconstructed videos, respectively. 
To enable temporal modeling, we insert the proposed RTS modules into the diffusion UNet of both TRUs. 
In addition, the Fake TRU is equipped with LoRA layers to enhance its capacity for learning the synthesized video distribution. 

\section{Model Analysis}
\subsection{Unique design of RTS}
Unlike previous works, our Recurrent Temporal Shift (RTS) module is specifically designed with the following unique features:
(1) Dual-level temporal modeling: Following the standard architecture of diffusion models (i.e., convolution + attention), RTS integrates a RTS-Convolution Unit for local temporal continuity through convolution layers and a RTS-Attention Unit for global semantic alignment using the Mutual Self-Attention.
(2) High efficiency: RTS is lightweight, with channel-reducing/increasing layers to minimize overhead, making it practical within diffusion-based pipelines.

\subsection{Computational efficiency of TAI}
Here, we analyze it from two perspectives: 
(1) TAI under memory-limited vs. non-TAI under larger-memory conditions: 
While larger memory enables faster inference, our TAI is specifically designed for memory-limited conditions (e.g., 48GB). Without TAI, prior methods must split videos into short clips, which disrupts temporal continuity. In contrast, TAI allows processing up to 120 frames at 2K resolution. 
(2) TAI vs. non-TAI under memory-limited conditions: 
In Tab. 2, we compare our method (with TAI) with MGLD-VSR (without TAI), both using the same backbone. Our method is 35.7$\times$ faster at 720p and 76.4$\times$ faster at 2K resolution, benefiting not only from fewer sampling steps but also from the efficient TAI module.

\subsection{Difference between Real TRU and fake TRU}
The Real TRU and Fake TRU differ in both structure and training strategy. 
Structurally, both variants consist of a frozen SD model and trainable RTS modules; however, Fake TRU additionally includes a trainable LoRA module. 
During the training of the One-Step VSR model, all other components are frozen, and both Real TRU and Fake TRU take the reconstructed HR video as input for computing the SJD loss. 
When updating Real TRU specifically, its input is the ground-truth video, and the loss follows the standard denoising objective~\cite{rombach2022high}. 
In contrast, Fake TRU is updated using the reconstructed HR video as input, also supervised by the standard denoising objective~\cite{rombach2022high}.

\section{User Study}
To strengthen claims of visual realism of our method, we conducted a user study to assess the perceptual quality of real-world videos generated by various diffusion-based VSR methods, including Upscale-A-Video~\cite{zhou2024upscale}, MGLD-VSR~\cite{yang2025motion}, STAR~\cite{xie2025star}, and our proposed method. 
We recruited 20 participants, each of whom viewed 20 randomly selected video triplets. 
Each triplet consisted of the input low-resolution video, the output of one competing method, and the output of our method. 
The order of the compared outputs was randomized. Participants were asked to select the video that appeared visually more realistic and temporally consistent. 
The results are shown in Fig.~\ref{fig:user}. 
Our approach gains more votes in the competitions.

\section{More Visual Results}
\begin{figure*}[t]
    \setlength{\abovecaptionskip}{0pt}
    \setlength{\belowcaptionskip}{-1pt}
    \centering
    \includegraphics[width=0.87\linewidth]{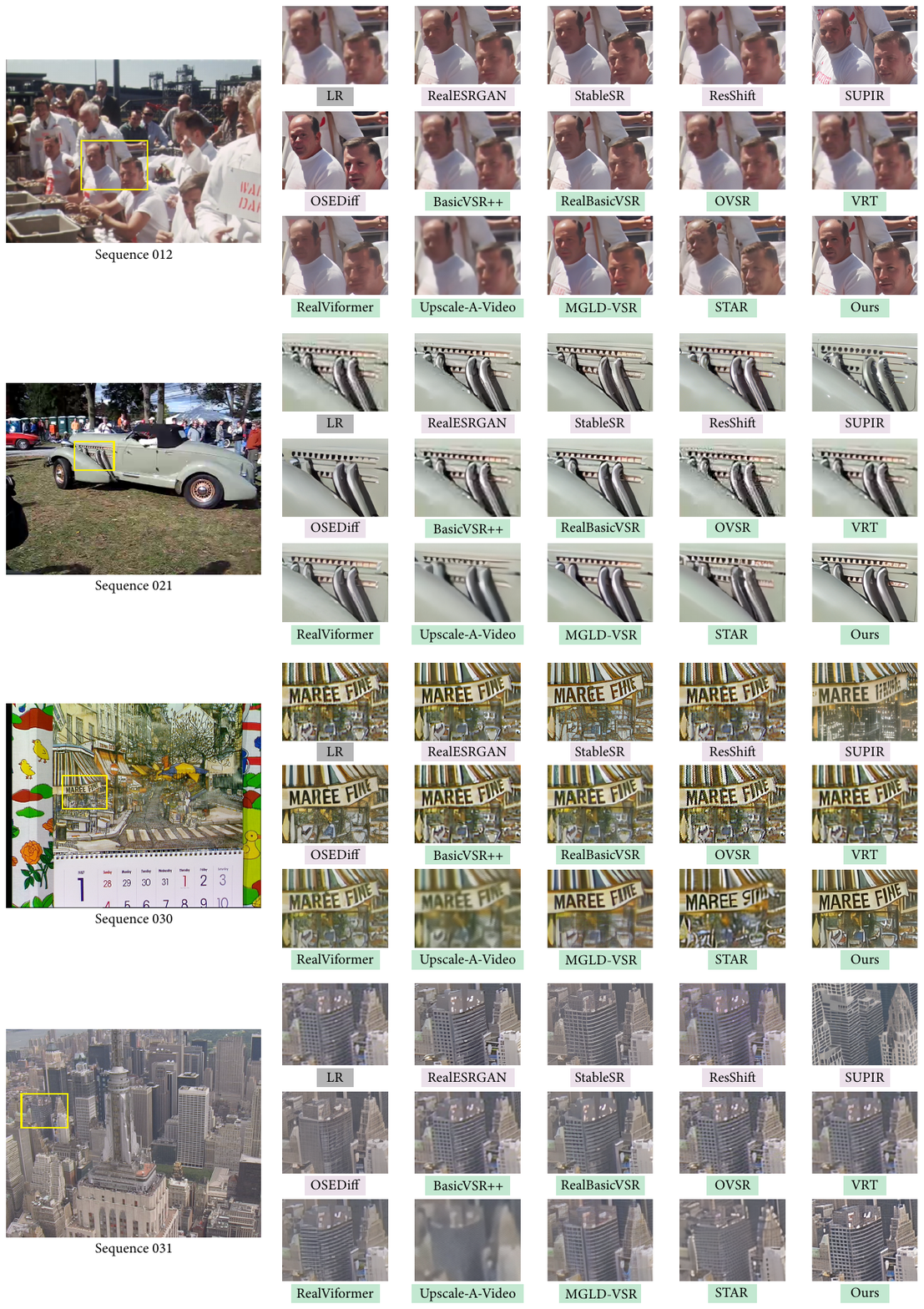}
    \caption{More visual comparisons both \colorbox{sisrcolor}{SISR} and \colorbox{vsrcolor}{VSR} methods on real-world low-quality videos from VideoLQ~\cite{chan2022investigating}dataset.}
    \label{fig:comparereal}
    \vspace{-1em}
\end{figure*}

\begin{figure*}[t]
    \setlength{\abovecaptionskip}{0pt}
    \setlength{\belowcaptionskip}{-1pt}
    \centering
    \includegraphics[width=0.87\linewidth]{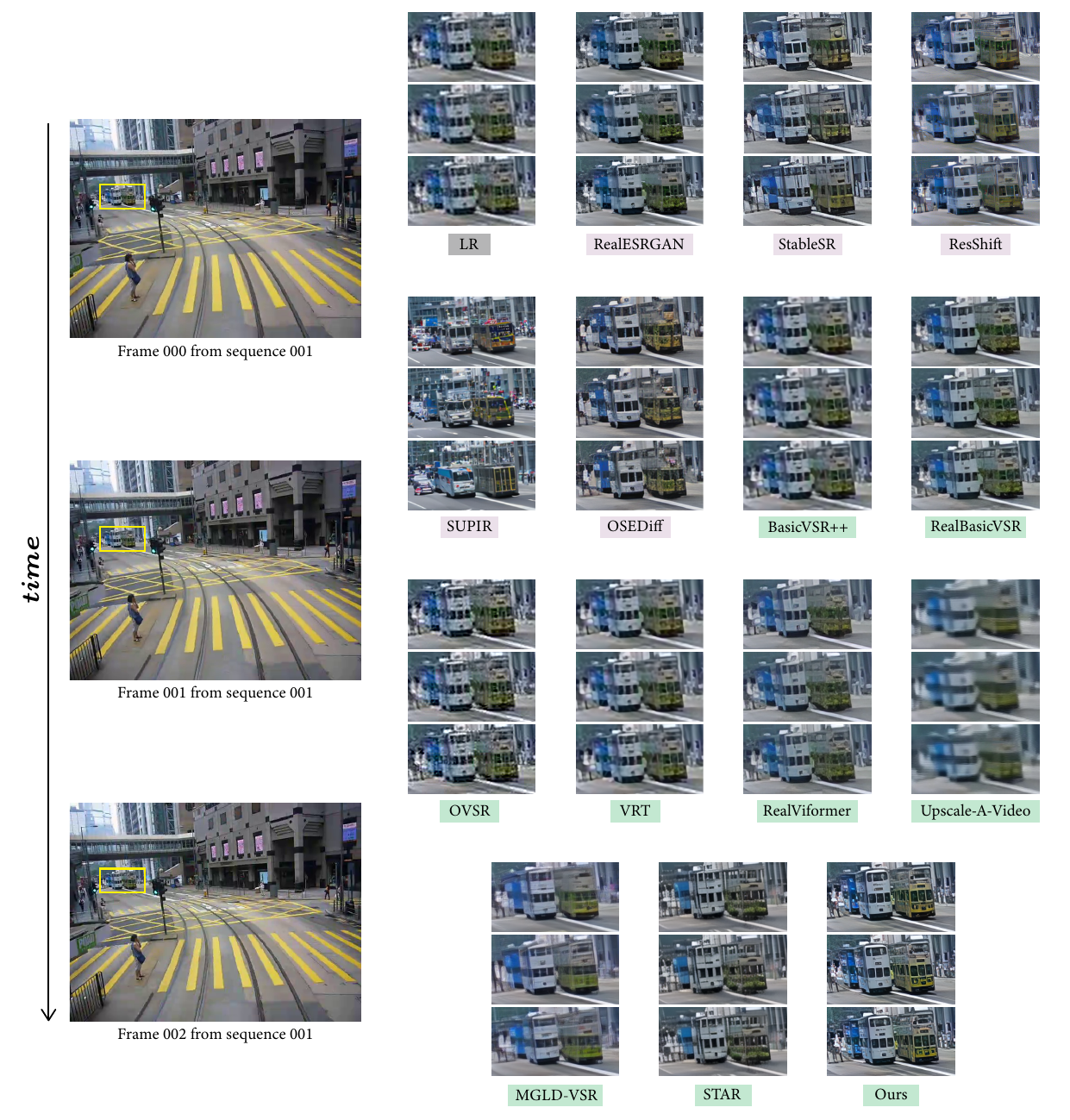}
    \caption{Temporal consistency analysis across both \colorbox{sisrcolor}{SISR} and \colorbox{vsrcolor}{VSR} methods.}
    \label{fig:comparetemportal}
    \vspace{-1em}
\end{figure*}

As shown in Fig.~\ref{fig:comparereal} and Fig.~\ref{fig:comparetemportal}, we present more visual comparisons and temporal consistency analysis with the state-of-the-art approaches. 
These results demonstrate that UltraVSR not only excels in perceptual quality but also maintains strong temporal coherence. 

\section{Video Demo}
We provide a \href{https://youtu.be/IqH3Y2-4hno}{\textcolor{red!80!black}{\textbf{video demo}}} to showcase more video comparison results with both diffusion-based SISR and VSR approaches.

\bibliographystyle{ACM-Reference-Format}
\bibliography{sample-base}